%% file: main.tex
\documentclass[10pt,twocolumn,letterpaper]{article}

\usepackage[pagenumbers]{cvpr} 

\input{preamble}

\definecolor{cvprblue}{rgb}{0.21,0.49,0.74}
\usepackage[pagebackref,breaklinks,colorlinks,allcolors=cvprblue]{hyperref}
\usepackage{multirow}
\usepackage{multicol}
\usepackage{algorithm}
\usepackage{algorithmic}
\def\confName{CVPR}
\def\confYear{2027}

\title{Score-based Black-box Attack via Adversarial Condition Guided Diffusion Model}

\author{%
  Yuhang Chen\textsuperscript{1}\quad
  Chih-Hong Cheng\textsuperscript{2,3}\quad
  Kurt Debattista\textsuperscript{1}\quad
  Siddartha Khastgir\textsuperscript{1}\quad
  Xingyu Zhao\textsuperscript{1,4}\textsuperscript{\dag}\\
  \textsuperscript{1}WMG, University of Warwick \quad
  \textsuperscript{2}University of Oldenburg \quad
  \textsuperscript{3}Chalmers University of Technology \\
  \textsuperscript{4}Wuhan University\\
  \small Email:\, \texttt{yuhang.chen.1@warwick.ac.uk} \quad \texttt{xingyu.zhao@whu.edu.cn} 
}

\begin{document}
\maketitle
\input{sec/0_abstract}    
\input{sec/1_intro}

\input{sec/2_preliminaries}

\input{sec/3_score}

\input{sec/4_advc}

\input{sec/5_experiment}

\input{sec/6_conclusion}
{
    \input{sec/7_llm_statement}
    \small
    \bibliographystyle{ieeenat_fullname}
    \bibliography{main}
}

\appendix
\input{sec/X_suppl}

\end{document}

%% file: preamble.tex
\usepackage{microtype}

\renewcommand{\paragraph}[1]{\vspace{.5em}\noindent\textbf{#1.}}



%% file: sec/0_abstract.tex
\begin{abstract}
Generative models enable unrestricted adversarial attacks that aim to fool classifiers while preserving the natural appearance and semantics of the original image. However, existing diffusion-based black-box attacks largely rely on transferability from surrogates, making their effectiveness sensitive to differences between surrogate and victim. Improving transferability can also distort semantic content and degrade image quality. Moreover, retaining the original prompt to preserve fidelity leaves the diffusion model’s conditional prior underused for adversarial guidance. We address these limitations by adapting score-based black-box attacks to reverse diffusion, using the victim’s output probabilities to guide latent updates without accessing its gradients. We further introduce adaptive-target guidance, which uses the victim’s current runner-up class as an adversarial condition to steer denoising towards the decision boundary. Because such guidance can introduce unwanted semantic changes, we also propose target-search guidance, which searches for a small perturbation of the original condition embedding using a score-based attack. This achieves comparable attack effectiveness with better image quality than runner-up guidance. Experiments on the ImageNet-Compatible dataset show that our framework outperforms transfer-based baselines in both attack effectiveness and image quality.
\end{abstract}

%% file: sec/1_intro.tex
\section{Introduction}
\label{sec:intro}

Despite their remarkable performance in visual recognition, deep neural networks (DNNs) are vulnerable to imperceptible perturbations on benign inputs that induce erroneous predictions \cite{DBLP:journals/corr/SzegedyZSBEGF13}, which pose severe security risks to safety-critical applications \cite{DBLP:journals/access/AkhtarM18}. Many attacks \cite{DBLP:conf/sp/Carlini017, DBLP:journals/corr/GoodfellowSS14, DBLP:conf/iclr/MadryMSTV18, DBLP:conf/eurosp/PapernotMJFCS16, DBLP:conf/cvpr/Moosavi-Dezfooli16, DBLP:conf/iclr/KurakinGB17a} optimize such perturbations to deceive perception models while preserving visual fidelity. Unlike classical attacks under $\ell_p$-norm constraints, \citet{DBLP:conf/nips/SongSKE18} introduced \textit{unrestricted adversarial examples} (UAEs), which synthesize semantically plausible inputs with generative priors~\cite{DBLP:conf/icml/OdenaOS17} to match human perception and evade defenses~\cite{DBLP:conf/icml/ZhangYJXGJ19, shi2021online}. It was later extended to non-$\ell_p$ perturbations, including spatial displacement~\cite{DBLP:conf/iclr/XiaoZ0HLS18}, texture manipulation~\cite{DBLP:conf/iclr/BhattadCLLF20} and color shifts~\cite{DBLP:conf/iclr/BhattadCLLF20, DBLP:conf/cvpr/HosseiniP18, DBLP:conf/cvpr/ShamsabadiSC20, DBLP:conf/bmvc/ZhaoLL20, DBLP:conf/nips/YuanZGCS22}. Diffusion models~\cite{DBLP:conf/nips/HoJA20, DBLP:conf/cvpr/RombachBLEO22} further enable photorealistic UAEs by perturbing the denoising trajectory~\cite{DBLP:conf/eccv/DaiLX24}. However, these methods assume white-box access to the victim, which rarely holds in practice and motivates black-box attacks~\cite{DBLP:conf/ccs/PapernotMGJCS17}.

\begin{figure}[t]
  \centering
  \includegraphics[width=\linewidth]{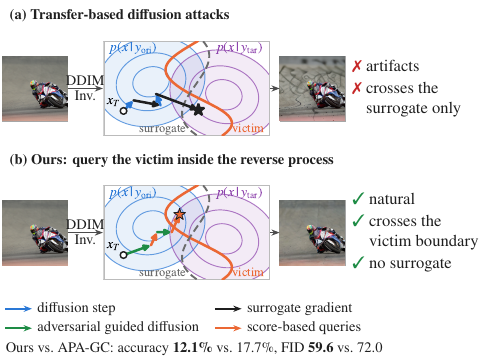}
  \caption{Transfer-based diffusion attacks versus ours. (a) The surrogate gradient pushes the sample towards $p(x\,|\,y_{\mathrm{tar}})$, yet it stays on the original side of the victim boundary. (b) Adversarial guidance and score-based queries move it across the victim boundary while it stays close to $p(x\,|\,y_{\mathrm{ori}})$.}
  \label{fig:teaser}

  \vspace{-2mm}
\end{figure}

In black-box settings, recent diffusion-based attacks largely follow the \textit{transfer-based} paradigm, crafting adversarial examples on white-box surrogate models with specialized objectives \cite{DBLP:journals/pami/ChenCCZZS25, DBLP:conf/nips/ChenLWJDZ23, DBLP:conf/nips/JiangCGLFGTLZ25, DBLP:conf/iclr/CollinsVFM26}. Their efficacy, however, fluctuates with the architectural alignment between the surrogate and the victim. Optimizing UAEs for high transferability also introduces visual semantics of the target class, compromising stealthiness \cite{DBLP:conf/iclr/ZhangDMK26}. Furthermore, to preserve visual fidelity, existing methods restrict the condition to a null token or the original label, and thus do not exploit the conditional prior of the diffusion model to steer the reverse process towards the decision boundary.

In this work, we propose, to the best of our knowledge, the first framework that integrates score-based black-box attacks into the reverse diffusion process, which avoids the surrogate mismatch and semantic distortion of transfer-based attacks by crafting UAEs tailored to each victim. By querying the scores of the victim instead of a surrogate, we align the perturbation with the decision boundary of the victim while keeping the image natural, as Figure~\ref{fig:teaser} illustrates. Specifically, we adapt six score-based attacks~\cite{DBLP:conf/icml/GuoGYWW19, DBLP:conf/eccv/AndriushchenkoC20, DBLP:conf/icml/IlyasEAL18, DBLP:conf/iclr/IlyasEM19, DBLP:conf/iclr/LiuCCH19, DBLP:conf/iclr/Al-DujailiO20} from the pixel space to the diffusion latent space, and study the latent bound, the number of inverted steps and the query budget per step.

Building on this framework, we develop two alternative algorithmic variants that exploit the diffusion model’s conditional prior to address different limitations of adversarial sampling. Both derive adversarial conditions from victim queries alone and retain the same score-based latent attack: the selected guidance modifies denoising before each latent update. Adaptive-target guidance (ATG) strengthens the attack under tight query budgets by adding guidance from the victim’s current runner-up class to the original reconstruction guidance. Selecting this class requires one extra victim query per step, while the guidance itself is supplied by the diffusion model. However, guidance towards a competing class can introduce unwanted semantic changes. To limit these changes, target-search guidance (TSG) instead searches for a small perturbation of the original prompt embedding using a score-based attack. This strengthens the attack while largely preserving the image quality of the unguided framework, at the cost of additional condition-search queries.

The main contributions of this paper are as follows:

\begin{itemize}

\item We adapt score-based attacks to reverse diffusion to craft UAEs for each victim without a surrogate. Across eleven victims, our unguided strategy achieves 12.1\% average accuracy and 59.6 FID with 720 queries per image, versus 17.7\% and 72.0 for the strongest transfer-based baseline.

\item We introduce adaptive-target guidance (ATG) and target-search guidance (TSG). ATG reduces average accuracy by 4.1--15.3 percentage points across six attacks with one extra query per step. At $Q=60$, TSG reduces accuracy by 5.9--24.6 points while increasing FID by at most 0.7.

\item We evaluate four adversarially trained models, where our strongest variants outperform transfer-based baselines in both accuracy and FID. On ResNet-50, our latent attack also outperforms the decision-based Latent-HSJA even when the latter receives ten times our query budget.

\end{itemize}

%% file: sec/2_preliminaries.tex
\section{Related Work}
\label{sec:related}

\paragraph{White-box diffusion attacks}
Unrestricted adversarial examples~\cite{DBLP:conf/nips/SongSKE18} replace the $L_p$-norm constraint with the requirement that the image remains natural and preserves its semantics, and diffusion models have become the main prior for such examples. AdvDiffuser~\cite{DBLP:conf/iccv/ChenGZY023} and DiffPGD~\cite{DBLP:conf/nips/XueA0C23} apply gradient-based perturbations during denoising, AdvDiff~\cite{DBLP:conf/eccv/DaiLX24} adds adversarial classifier guidance to the reverse process, and Zhang \etal~\cite{DBLP:conf/nips/0001ZW24} sample from the product of a victim distribution and a distance distribution. These attacks require victim gradients, whereas our framework uses queried probabilities to guide latent updates during reverse diffusion.

\paragraph{Transfer-based attacks}
Without access to the victim, transfer-based attacks craft adversarial examples on a white-box surrogate and rely on their transferability~\cite{DBLP:conf/ccs/PapernotMGJCS17}. Perturbation-based attacks such as MI-FGSM~\cite{DBLP:conf/cvpr/DongLPS0HL18}, DI-FGSM~\cite{DBLP:conf/cvpr/XieZZBWRY19}, TI-FGSM~\cite{DBLP:conf/cvpr/DongPSZ19}, PI-FGSM~\cite{DBLP:conf/eccv/GaoZSLS20} and S$^2$I-FGSM~\cite{DBLP:conf/eccv/LongZZGLZS22} improve transferability with momentum, input transformations and spectrum augmentation. Unrestricted attacks add dedicated designs for the same goal: NCF~\cite{DBLP:conf/nips/YuanZGCS22} searches natural color distributions with neighborhood search and initialization reset, DiffAttack~\cite{DBLP:journals/pami/ChenCCZZS25} further deceives the diffusion model by distracting its attention, ACA~\cite{DBLP:conf/nips/ChenLWJDZ23} optimizes the latent on the natural-image manifold, and APA~\cite{DBLP:conf/nips/JiangCGLFGTLZ25} aggregates gradients over diffusion steps. NatADiff~\cite{DBLP:conf/iclr/CollinsVFM26} combines augmented classifier guidance with time-travel sampling, guided by a prompt that names both the true class and a pre-selected target class, and synthesizes each sample from noise instead of perturbing a given image. Their effectiveness depends on surrogate--victim alignment, and adversarial training can reduce transferability~\cite{DBLP:conf/iclr/TramerKPGBM18}. Our framework instead uses direct victim queries to guide both latent updates and adversarial conditioning.

\paragraph{Query-based attacks}
Query-based attacks instead observe the outputs of the victim. Decision-based attacks use only the predicted label~\cite{DBLP:conf/iclr/BrendelRB18, DBLP:conf/sp/ChenJW20}, whereas score-based attacks use the output probabilities to estimate the gradient or its sign, as in NES~\cite{DBLP:conf/icml/IlyasEAL18}, Bandits~\cite{DBLP:conf/iclr/IlyasEM19}, ZOsignSGD~\cite{DBLP:conf/iclr/LiuCCH19} and SignHunter~\cite{DBLP:conf/iclr/Al-DujailiO20}, or to accept sampled updates, as in SimBA~\cite{DBLP:conf/icml/GuoGYWW19} and Square~\cite{DBLP:conf/eccv/AndriushchenkoC20}. These methods typically perturb pixels within an $L_p$ ball and can require thousands of queries per image~\cite{DBLP:journals/pami/ZhengYZCW25}. A few attacks combine queries with a generative model. Latent-HSJA~\cite{DBLP:conf/eccv/NaJK22} runs HopSkipJumpAttack in the latent space of a GAN, and AdvDiffEdit~\cite{wang2026decision} edits images with a diffusion model under accept or reject feedback, both observing only the decision. AdvFlow~\cite{DBLP:conf/nips/DolatabadiEL20} applies NES in the latent space of a normalizing flow, but maps each latent to an image in one pass and projects it into an $L_p$ ball. Our framework instead uses victim probabilities to update diffusion latents across successive denoising steps, without imposing a final pixel-space $L_p$ constraint.

%% file: sec/3_score.tex
\section{Latent Score-Based Black-Box Adversarial Sampling}
\label{sec:method}

\begin{figure*}[t]
  \centering
  \includegraphics[width=\textwidth]{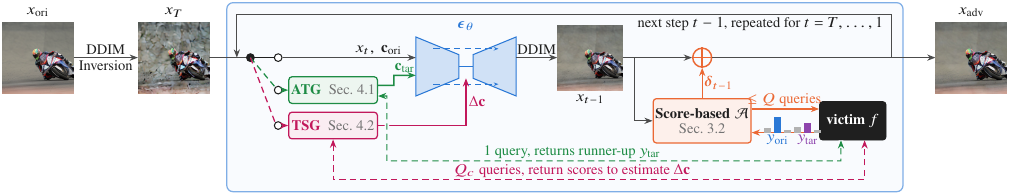}
  \caption{Overview of latent score-based adversarial sampling. The clean image is inverted to $x_T$ and denoised step by step. At each step, the switch keeps the original condition $\mathbf{c}_{\mathrm{ori}}$ or routes $x_t$ to ATG or TSG, and the score-based attack updates $x_{t-1}$ from the scores of at most $Q$ queries to the victim. Thumbnails are decoded latents of Latent-NES against ResNet-50.}
  \label{fig:framework}
\end{figure*}

\subsection{Preliminaries}
\label{sec:prelim}

\paragraph{Adversarial examples}
Let $f$ be a classifier over images $x\in[0,1]^n$ with label space $\mathcal{Y}$ and class probabilities $p_f(y| x)$. Given an image $x_{\mathrm{ori}}$ of class $y_{\mathrm{ori}}$, the targeted attack objective for an incorrect class \(y_{\mathrm{tar}}\) is~\cite{DBLP:journals/corr/SzegedyZSBEGF13}:
\begin{equation}
  \min_{x_{\mathrm{adv}}}\ d(x_{\mathrm{ori}}, x_{\mathrm{adv}}) \quad \text{s.t.}\quad f(x_{\mathrm{adv}})=y_{\mathrm{tar}},\ \ x_{\mathrm{adv}}\in[0,1]^n,
  \label{eq:adv-def}
\end{equation}
where $d$ is a distance between images. As this problem is hard to solve, it is relaxed to~\cite{DBLP:journals/corr/SzegedyZSBEGF13,DBLP:conf/sp/Carlini017}
\begin{equation}
  \min_{x_{\mathrm{adv}}\in[0,1]^n}\ c_1\, d(x_{\mathrm{ori}}, x_{\mathrm{adv}}) + c_2\, \mathcal{L}(x_{\mathrm{adv}}, y_{\mathrm{tar}}),
  \label{eq:adv-relax}
\end{equation}
where $\mathcal{L}$ is a classification loss such as the cross-entropy $\mathcal{L}(x, y)=-\log p_f(y| x)$. Since we do not attack towards a pre-specified class, $y_{\mathrm{tar}}\in\mathcal{Y}\setminus\{y_{\mathrm{ori}}\}$ is any incorrect class.

\paragraph{Adversarial distribution}
Zhang \etal~\cite{DBLP:conf/nips/0001ZW24} view adversarial examples as samples from an adversarial distribution that is a product of experts~\cite{DBLP:journals/neco/Hinton02},
\begin{equation}
  p_{\mathrm{adv}}(x| x_{\mathrm{ori}}, y_{\mathrm{tar}}) \propto p_{\mathrm{vic}}(x| y_{\mathrm{tar}})\, p_{\mathrm{dis}}(x| x_{\mathrm{ori}}),
  \label{eq:padv}
\end{equation}
where the victim distribution $p_{\mathrm{vic}}(x| y_{\mathrm{tar}})\propto e^{-c\,\mathcal{L}(x, y_{\mathrm{tar}})}$ favors inputs classified as $y_{\mathrm{tar}}$ and the distance distribution $p_{\mathrm{dis}}(x| x_{\mathrm{ori}})\propto e^{-d(x_{\mathrm{ori}}, x)}$ favors inputs close to $x_{\mathrm{ori}}$. With these two experts, Langevin sampling from $p_{\mathrm{adv}}$ provides a sampling-based approach to the objective in Eq.~\eqref{eq:adv-relax}. Replacing $p_{\mathrm{dis}}$ by a diffusion model that reconstructs $x_{\mathrm{ori}}$ with Gaussian steps $p_\theta(x_{t-1}| x_t)$ of mean $\mu_\theta(x_t)$ and covariance $\Sigma_\theta(x_t)$, and distributing the victim distribution over the $T$ denoising steps, they approximate Eq.~\eqref{eq:padv} as
\begin{equation}
\begin{split}
  &p_{\mathrm{adv}}(x_0| x_{\mathrm{ori}}, y_{\mathrm{tar}}) \\
  &\ \propto \int p(x_T)\prod_{t=1}^{T} p_{\mathrm{vic}}(\hat{x}_{0|t-1}| y_{\mathrm{tar}})^{\frac{1}{T}}\, p_\theta(x_{t-1}| x_t)\, \mathrm{d}x_{1:T},
\end{split}
  \label{eq:padv-diffusion}
\end{equation}
where Tweedie's estimate of the clean image~\cite{efron2011tweedie} is $\hat{x}_{0|t}=\bigl(x_t-\sqrt{1-\bar{\alpha}_t}\,\boldsymbol{\epsilon}_\theta(x_t)\bigr)/\sqrt{\bar{\alpha}_t}$. After normalization, each factor admits a first-order Gaussian approximation~\cite{DBLP:conf/nips/0001ZW24},
\begin{equation}
\begin{split}
  &p_{\mathrm{vic}}(\hat{x}_{0|t-1}| y_{\mathrm{tar}})^{\frac{1}{T}}\, p_\theta(x_{t-1}| x_t) \\
  &\ \approx \mathcal{N}\bigl(x_{t-1};\, \mu_\theta(x_t)+\tfrac{c}{T}\Sigma_\theta(x_t)\,\mathbf{g}_t,\ \Sigma_\theta(x_t)\bigr),
\end{split}
\label{eq:padv-step}
\end{equation}
with $\mathbf{g}_t = -\nabla_{x_{t-1}}\mathcal{L}(\hat{x}_{0|t-1}, y_{\mathrm{tar}})\big|_{x_{t-1}=\mu_\theta(x_t)}$, so that an adversarial example is drawn by the standard reverse process with the gradient of the victim loss added to the mean of every step. White-box diffusion-based attacks obtain $\mathbf{g}_t$ by back-propagating through $f$.

\subsection{Score-Based Gradient Estimation}
\label{sec:score}

In the score-based setting, $f$ only returns $p_f(\cdot| x)$, so $\mathbf{g}_t$ in Eq.~\eqref{eq:padv-step} cannot be computed. We estimate it with score-based attacks that maximize the log-probability of the target class,
\begin{equation}
  S(x) = \log p_f(y_{\mathrm{tar}}| x) = -\mathcal{L}(x, y_{\mathrm{tar}}),
  \label{eq:score-obj}
\end{equation}
whose gradient corresponds to $\mathbf{g}_t$. As no target class is specified, $y_{\mathrm{tar}}$ is selected from the prediction of the victim at every query,
\begin{equation}
  y_{\mathrm{tar}} = \begin{cases}
    \arg\max_{y\neq y_{\mathrm{ori}}} p_f(y| x), & \text{if } f(x) = y_{\mathrm{ori}},\\[2pt]
    f(x), & \text{otherwise},
  \end{cases}
  \label{eq:ytar}
\end{equation}
We adapt six score-based attacks, each of which returns an update $\boldsymbol{\delta}$ with $\|\boldsymbol{\delta}\|_\infty\le\varepsilon$ from at most $Q$ queries, where $\Pi_\varepsilon$ projects onto the \(L_\infty\) ball of radius $\varepsilon$ centered at the input to the current attack call.
NES~\cite{DBLP:conf/icml/IlyasEAL18} estimates the gradient with $q$ antithetic Gaussian directions, and ZOsignSGD~\cite{DBLP:conf/iclr/LiuCCH19} with $q$ forward differences,
\begin{equation}
\begin{split}
  \hat{\mathbf{g}}_{\mathrm{NES}} &= \sum_{k=1}^{q}\frac{S(x+\eta\mathbf{u}_k)-S(x-\eta\mathbf{u}_k)}{2q\eta}\,\mathbf{u}_k, \\
  \hat{\mathbf{g}}_{\mathrm{ZO}} &= \sum_{k=1}^{q}\frac{S(x+\eta\mathbf{u}_k)-S(x)}{q\eta}\,\mathbf{u}_k,
\end{split}
  \label{eq:nes}
\end{equation}
with $\mathbf{u}_k\sim\mathcal{N}(0,\mathbf{I})$, and both take the signed step $x\leftarrow\Pi_\varepsilon(x+\alpha\,\mathrm{sign}(\hat{\mathbf{g}}))$. Bandits~\cite{DBLP:conf/iclr/IlyasEM19} maintains a prior $\mathbf{v}$ of the gradient that is refined across iterations,
\begin{equation}
\begin{split}
  \mathbf{v}&\leftarrow\mathbf{v}+h\,\frac{S(x+\eta(\mathbf{v}+\zeta\mathbf{u}))-S(x+\eta(\mathbf{v}-\zeta\mathbf{u}))}{\eta\zeta}\,\mathbf{u}, \\
  x&\leftarrow\Pi_\varepsilon\bigl(x+\alpha\,\mathrm{sign}(\mathbf{v})\bigr).
\end{split}
  \label{eq:bandits}
\end{equation}
SignHunter~\cite{DBLP:conf/iclr/Al-DujailiO20} estimates the sign vector $\mathbf{s}\in\{-1,1\}^{m}$ of the gradient directly, flipping the signs of coordinate blocks from coarse to fine and keeping a flip whenever it increases $S(x+\varepsilon\mathbf{s})$.

Random-search attacks instead accept sampled updates. SimBA~\cite{DBLP:conf/icml/GuoGYWW19} samples a direction $\mathbf{e}$ from an orthonormal basis and accepts $x\pm\alpha\mathbf{e}$ whenever it increases $S$, and Square Attack~\cite{DBLP:conf/eccv/AndriushchenkoC20} sets a random square window to $\pm\varepsilon$ per channel, a vertex of the $L_\infty$ ball, accepting the proposal whenever it increases $S$.

\subsection{Adver. Sampling with Estimated Gradients}
\label{sec:framework}

Motivated by Eq.~\eqref{eq:padv-step}, we combine a latent diffusion model~\cite{DBLP:conf/cvpr/RombachBLEO22} and DDIM sampling ~\cite{DBLP:conf/iclr/SongME21} with score-based adversarial updates, as demonstrated in Figure~\ref{fig:framework}. We write $x_t$ for its latents and omit the encoder and decoder, so that $x_0$ is the latent of $x_{\mathrm{ori}}$ and the victim receives decoded images. Following DiffAttack~\cite{DBLP:journals/pami/ChenCCZZS25}, we obtain $x_T$ by DDIM Inversion of $x_0$ over the last $T$ steps and optimize the unconditional embedding to reconstruct $x_{\mathrm{ori}}$. Each reverse step uses classifier-free guidance~\cite{DBLP:journals/corr/abs-2207-12598} for the class prompt $\mathbf{c}_{\mathrm{ori}}$ with scale $w$,
\begin{equation}
  \tilde{\boldsymbol{\epsilon}}_\theta(x_t) = \boldsymbol{\epsilon}_\theta(x_t,t,\varnothing)+w\bigl[\boldsymbol{\epsilon}_\theta(x_t,t,\mathbf{c}_{\mathrm{ori}})-\boldsymbol{\epsilon}_\theta(x_t,t,\varnothing)\bigr],
  \label{eq:cfg}
\end{equation}
followed by the deterministic DDIM update
\begin{equation}
  x_{t-1} = \sqrt{\bar{\alpha}_{t-1}}\,\hat{x}_{0|t}+\sqrt{1-\bar{\alpha}_{t-1}}\,\tilde{\boldsymbol{\epsilon}}_\theta(x_t),
  \label{eq:ddim-step}
\end{equation}
with $\hat{x}_{0|t}$ computed from $\tilde{\boldsymbol{\epsilon}}_\theta$. As DDIM is deterministic, $x_{t-1}$ equals the mean $\mu_\theta(x_t)$ at which Eq.~\eqref{eq:padv-step} evaluates $\mathbf{g}_t$. We therefore estimate the gradient at $x_{t-1}$ with a score-based attack $\mathcal{A}$ of at most $Q$ queries and add its update,
\begin{equation}
  x_{t-1} \leftarrow x_{t-1}+\boldsymbol{\delta}_{t-1}, \quad \boldsymbol{\delta}_{t-1}=\mathcal{A}(x_{t-1}, S, \varepsilon, Q),
  \label{eq:perturb}
\end{equation}
where $\varepsilon$ bounds the perturbation returned by each attack call. In the final denoising steps, we query the victim on decoded intermediate latents as a cheaper surrogate for evaluation through the clean estimate. Each attack call stops once the decoded candidate is misclassified, denoising then continues. The inverted $x_T$ is updated in the same way, and sampling and updating alternate until $x_0$, which is decoded into $x_{\mathrm{adv}}$.
\label{sec:parameters}The hyperparameters are $T$, $\varepsilon$ and $Q$, and an image costs at most $(T+1)Q$ queries. Attack-specific settings are in the supplementary material.

%% file: sec/4_advc.tex
\section{Adversarial Sampling with Condition Guidance}
\label{sec:advcond}

Under a tight per-step query budget $Q$, score-based latent updates receive limited victim feedback. Meanwhile, denoising under the original condition favors reconstruction without explicitly promoting misclassification. We therefore adapt the condition to make denoising support the adversarial objective.

Motivated by the distributional perspective on adversarial attacks~\cite{DBLP:journals/tip/ZhuCLC00ZCH22}, we use the diffusion model’s conditional prior to guide denoising toward a competing class. This provides a class-dependent direction that complements the victim-based latent updates. We evaluate its contribution to attack effectiveness in Sec.~\ref{sec:guidance}.

\subsection{Adaptive-Target Guidance}
\label{sec:atg}
Inspired by classifier-free guidance~\cite{DBLP:journals/corr/abs-2207-12598}, adaptive-target guidance (ATG) augments the original reconstruction guidance with a competing-class condition. Under ideal conditional and unconditional scores, Bayes’ rule gives the implicit classifier direction:
\begin{equation}
\begin{split}
  \nabla_{x_t}\log p_\theta(y_{\mathrm{tar}}| x_t) &= \nabla_{x_t}\log\frac{p_\theta(x_t| y_{\mathrm{tar}})}{p_\theta(x_t)} \\
  &\approx -\frac{\boldsymbol{\epsilon}_\theta(x_t,t,\mathbf{c}_{\mathrm{tar}})-\boldsymbol{\epsilon}_\theta(x_t,t,\varnothing)}{\sqrt{1-\bar{\alpha}_t}}.
\end{split}
  \label{eq:implicit-classifier}
\end{equation}
Since the unconditional embedding is optimized for reconstruction in Sec.~\ref{sec:framework}, Eq.~\eqref{eq:implicit-classifier} provides an idealized motivation for the implemented guidance.
Motivated by Eq.~\eqref{eq:implicit-classifier}, we add the competing-class noise difference to the reconstruction-guided prediction with weight \(s\). The resulting noise prediction is used in the DDIM update:
\begin{equation}
\begin{split}
  \tilde{\boldsymbol{\epsilon}}_\theta(x_t) = \ &\boldsymbol{\epsilon}_\theta(x_t,t,\varnothing) + w\bigl[\boldsymbol{\epsilon}_\theta(x_t,t,\mathbf{c}_{\mathrm{ori}})-\boldsymbol{\epsilon}_\theta(x_t,t,\varnothing)\bigr] \\
  &+ s\bigl[\boldsymbol{\epsilon}_\theta(x_t,t,\mathbf{c}_{\mathrm{tar}})-\boldsymbol{\epsilon}_\theta(x_t,t,\varnothing)\bigr],
\end{split}
  \label{eq:atg}
\end{equation}

The guided DDIM step is followed by the score-based latent update in Eq.~\eqref{eq:perturb}. At each step, one victim query selects $y_{\mathrm{tar}}$ according to Eq.~\eqref{eq:ytar}, the highest-probability class other than $y_{\mathrm{ori}}$.

\subsection{Target-Search Guidance}
\label{sec:tsg-method}

At large guidance scales, ATG can introduce features of the competing class. Target-search guidance (TSG) instead searches near the original prompt embedding to improve the attack score while limiting semantic changes. Specifically, it seeks a condition within $\|\mathbf{c}-\mathbf{c}_{\mathrm{ori}}\|_\infty\leq\eta_c$ that increases $S(\hat{x}_{0|t}(\mathbf{c}))$, where $\hat{x}_{0|t}(\mathbf{c})$ is the clean latent estimate at step $t$ under condition $\mathbf{c}$, with decoding implicit in $S$. We use ZOsignSGD to estimate the condition gradient from $Q_c$ forward differences:
\begin{equation}
  \hat{\mathbf{g}}_c = \sum_{k=1}^{Q_c}\frac{S\bigl(\hat{x}_{0|t}(\mathbf{c}_{\mathrm{ori}}+\eta_c\mathbf{u}_k)\bigr)-S\bigl(\hat{x}_{0|t}(\mathbf{c}_{\mathrm{ori}})\bigr)}{\eta_c}\,\mathbf{u}_k,
  \label{eq:tsg-grad}
\end{equation}
where $\mathbf{u}_k\sim\mathcal{N}(0,\mathbf{I})$. Using $\eta_c$ as both the finite-difference scale and the embedding radius, we take one signed ascent step:
\begin{equation}
  \mathbf{c}_{\mathrm{adv}} = \mathbf{c}_{\mathrm{ori}} + \eta_c\,\mathrm{sign}(\hat{\mathbf{g}}_c).
  \label{eq:tsg-update}
\end{equation}
The reverse step uses $\mathbf{c}_{\mathrm{adv}}$ in place of $\mathbf{c}_{\mathrm{ori}}$ in Eq.~\eqref{eq:cfg}, followed by the latent update in Eq.~\eqref{eq:perturb}.

\subsection{Combined Algorithm and Complexity}
\label{sec:complexity}

\begin{algorithm}[t]
\caption{Latent score-based adversarial sampling}
\label{alg:sampling}
\renewcommand{\algorithmicrequire}{\textbf{Input:}}
\renewcommand{\algorithmicensure}{\textbf{Output:}}
\begin{algorithmic}[1]
\REQUIRE image $x_{\mathrm{ori}}$ with label $y_{\mathrm{ori}}$, victim $f$, attack $\mathcal{A}$, $T$, $\varepsilon$, $Q$, $w$, ATG scale $s$ or TSG budget $Q_c$
\ENSURE adversarial example $x_{\mathrm{adv}}$
\STATE $x_T \leftarrow$ DDIM Inversion of $x_0$, the latent of $x_{\mathrm{ori}}$
\STATE $x_T \leftarrow x_T + \mathcal{A}(x_T, S, \varepsilon, Q)$
\FOR{$t = T, \ldots, 1$}
\IF{ATG}
\STATE $y_{\mathrm{tar}} \leftarrow$ Eq.~\eqref{eq:ytar} at $x_t$, \ $\tilde{\boldsymbol{\epsilon}}_\theta(x_t) \leftarrow$ Eq.~\eqref{eq:atg}
\ELSIF{TSG}
\STATE $\mathbf{c}_{\mathrm{adv}} \leftarrow$ Eq.~\eqref{eq:tsg-update}, \ $\tilde{\boldsymbol{\epsilon}}_\theta(x_t) \leftarrow$ Eq.~\eqref{eq:cfg} with $\mathbf{c}_{\mathrm{adv}}$
\ELSE
\STATE $\tilde{\boldsymbol{\epsilon}}_\theta(x_t) \leftarrow$ Eq.~\eqref{eq:cfg}
\ENDIF
\STATE $x_{t-1} \leftarrow$ Eq.~\eqref{eq:ddim-step}
\STATE $x_{t-1} \leftarrow x_{t-1} + \mathcal{A}(x_{t-1}, S, \varepsilon, Q)$
\ENDFOR
\STATE \textbf{return} the decoded $x_0$ as $x_{\mathrm{adv}}$
\end{algorithmic}
\end{algorithm}

Let \(C_f\) denote the cost of decoding and one victim forward pass, and \(C_\epsilon\) the cost of one noise prediction. Excluding inversion and reconstruction-embedding optimization, the unguided attack uses \(2T\) noise predictions and at most \((T+1)Q\) victim queries. ATG adds \(T\) noise predictions and at most \(T\) victim queries. TSG adds \(T(Q_c+1)\) victim queries and \(O(TQ_c)\) noise predictions. Its total time complexity is therefore
\[
O\!\left(T(Q+Q_c)C_f+T(1+Q_c)C_\epsilon\right).
\]
ATG requires little additional computation, whereas TSG spends additional queries on condition search to limit the image changes associated with explicit competing-class guidance.

%% file: sec/5_experiment.tex
\section{Experiments}

\begin{table*}[!t]
  \centering
  \caption{\label{tab:main}%
    Top-1 accuracy (\%, $\downarrow$) of the victim models and FID ($\downarrow$) on the ImageNet-Compatible Dataset. Transfer-based attacks are averaged over all other models used as surrogates. Our attacks use $Q$ queries per active step without guidance (---), with ATG at scale $s$ (ATG $s$), or with TSG spending $Q_c$ extra queries per step (TSG $Q_c$). The best result in each column is shown in \textcolor{red}{\textbf{red}} and the second best is \underline{underlined}.
  }
  \resizebox{0.99\textwidth}{!}{
    \begin{tabular}{lccccccccccccccc}
    \toprule[1.5pt]
    \multirow{2}{*}{Attack} & \multirow{2}{*}{$Q$} & \multirow{2}{*}{Guidance} & \multicolumn{5}{c}{CNNs} & \multicolumn{4}{c}{Transformers} & \multicolumn{2}{c}{MLPs} & \multirow{2}{*}{Avg.} & \multirow{2}{*}{FID} \\
    \cmidrule(lr){4-8} \cmidrule(lr){9-12} \cmidrule(lr){13-14}
    & & & Res50 & VGG19 & MobV2 & IncV3 & ConvNeXt & ViT-B & Swin-B & DeiT-B & DeiT-S & Mixer-B & Mixer-L & & \\
    \midrule
    Clean & --- & --- & 92.7 & 88.7 & 86.9 & 80.5 & 97.0 & 93.7 & 95.9 & 94.5 & 94.0 & 82.5 & 76.5 & 89.4 & 57.8 \\
    \midrule
    PI-FGSM & --- & --- & 32.4 & 21.2 & 23.5 & 30.0 & 69.3 & 54.6 & 66.5 & 62.4 & 48.9 & 43.4 & 34.7 & 44.3 & 94.0 \\
    S$^2$I-FGSM & --- & --- & 37.2 & 30.5 & 28.0 & 33.2 & 57.1 & 54.0 & 49.1 & 54.1 & 44.3 & 38.4 & 34.7 & 41.9 & 77.2 \\
    NCF & --- & --- & 47.0 & 40.9 & 41.0 & 56.1 & 77.6 & 66.9 & 74.1 & 72.7 & 66.1 & 53.5 & 46.8 & 58.4 & 67.6 \\
    DiffAttack & --- & --- & 37.5 & 37.9 & 35.3 & 34.5 & 47.8 & 45.7 & 40.2 & 44.0 & 39.0 & 36.6 & 35.3 & 39.4 & 65.4 \\
    ACA & --- & --- & 38.5 & 35.7 & 34.2 & 32.8 & 47.7 & 44.7 & 45.8 & 43.2 & 42.3 & 31.7 & 28.4 & 38.6 & 64.8 \\
    APA-SG & --- & --- & 22.8 & 24.6 & 21.5 & 24.0 & 39.3 & 32.1 & 34.5 & 33.0 & 26.7 & 25.5 & 24.5 & 28.0 & 65.2 \\
    APA-GC & --- & --- & 12.8 & 14.5 & 11.9 & 12.4 & 24.0 & 21.0 & 21.2 & 22.5 & 17.0 & 17.4 & 19.5 & 17.7 & 72.0 \\
    NatADiff & --- & --- & 33.7 & 31.6 & 31.6 & 30.1 & 38.2 & 37.7 & 37.6 & 37.5 & 35.3 & 32.7 & 31.3 & 34.3 & 66.1 \\
    \midrule
    \multirow{7}{*}{\shortstack[l]{Latent-\\NES}} & \multirow{4}{*}{60} & --- & 22.4 & 19.8 & 18.8 & 12.4 & 40.5 & 28.1 & 36.2 & 32.9 & 27.6 & 15.0 & 12.7 & 24.2 & 59.6 \\
     &  & ATG 1.5 & 16.3 & 11.9 & 13.1 & 8.6 & 32.2 & 20.5 & 27.6 & 23.3 & 20.5 & 9.3 & 8.4 & 17.4 & 63.2 \\
     &  & ATG 2.5 & 11.3 & 9.3 & 10.9 & 6.9 & 24.3 & 15.7 & 20.7 & 20.2 & 16.8 & 7.3 & 6.5 & 13.6 & 67.1 \\
     &  & TSG 60 & 14.0 & 10.6 & 12.4 & 7.0 & 29.4 & 20.1 & 27.0 & 23.6 & 18.8 & 8.5 & 7.8 & 16.3 & 60.1 \\
    \cmidrule(lr){2-16}
     & \multirow{3}{*}{120} & --- & 11.5 & 9.6 & 8.8 & 5.2 & 23.1 & 13.5 & 21.6 & 19.1 & 15.0 & 5.5 & 5.7 & 12.6 & 59.6 \\
     &  & ATG 1.5 & 6.1 & 6.5 & 5.3 & 3.3 & 16.4 & 9.0 & 16.0 & 14.6 & \underline{8.8} & 5.3 & \underline{2.2} & 8.5 & 63.2 \\
     &  & ATG 2.5 & \underline{5.7} & \underline{4.4} & \underline{5.1} & \underline{3.0} & \underline{13.4} & 7.7 & \underline{12.3} & 10.7 & \textcolor{red}{\textbf{8.4}} & 2.9 & 3.2 & \underline{7.0} & 67.4 \\
    \midrule
    \multirow{7}{*}{\shortstack[l]{Latent-\\ZOsignSGD}} & \multirow{4}{*}{60} & --- & 21.1 & 16.8 & 19.3 & 10.8 & 34.3 & 24.7 & 33.2 & 28.4 & 25.6 & 10.6 & 9.0 & 21.3 & 59.4 \\
     &  & ATG 1.5 & 14.4 & 11.0 & 11.9 & 7.0 & 25.2 & 15.6 & 25.2 & 21.3 & 16.7 & 6.6 & 6.2 & 14.6 & 62.8 \\
     &  & ATG 2.5 & 10.3 & 8.7 & 10.8 & 6.1 & 20.6 & 13.0 & 19.5 & 16.7 & 13.8 & 4.8 & 4.6 & 11.7 & 67.5 \\
     &  & TSG 60 & 13.5 & 11.4 & 11.4 & 6.7 & 27.1 & 19.0 & 25.5 & 22.6 & 18.8 & 6.7 & 6.2 & 15.4 & 60.1 \\
    \cmidrule(lr){2-16}
     & \multirow{3}{*}{120} & --- & 10.3 & 9.2 & 9.7 & 5.3 & 22.4 & 13.1 & 22.4 & 18.9 & 13.2 & 4.6 & 4.5 & 12.1 & 59.6 \\
     &  & ATG 1.5 & 6.7 & 4.6 & 5.9 & 3.1 & 13.8 & 9.9 & 16.1 & 11.4 & 10.2 & 3.1 & 2.3 & 7.9 & 63.0 \\
     &  & ATG 2.5 & \textcolor{red}{\textbf{5.3}} & \textcolor{red}{\textbf{3.0}} & \textcolor{red}{\textbf{4.6}} & \textcolor{red}{\textbf{2.1}} & \textcolor{red}{\textbf{12.0}} & 7.3 & \textcolor{red}{\textbf{10.9}} & \textcolor{red}{\textbf{8.8}} & 9.0 & \textcolor{red}{\textbf{2.1}} & \textcolor{red}{\textbf{1.7}} & \textcolor{red}{\textbf{6.1}} & 67.4 \\
    \midrule
    \multirow{7}{*}{\shortstack[l]{Latent-\\SignHunter}} & \multirow{4}{*}{60} & --- & 19.6 & 19.9 & 17.1 & 17.9 & 37.9 & 25.0 & 35.9 & 27.4 & 28.2 & 10.7 & 8.2 & 22.5 & 62.7 \\
     &  & ATG 1.5 & 15.8 & 14.0 & 14.1 & 12.7 & 27.0 & 17.1 & 26.2 & 22.9 & 18.7 & 7.9 & 5.7 & 16.6 & 66.2 \\
     &  & ATG 2.5 & 12.0 & 10.7 & 10.8 & 8.9 & 22.2 & 14.5 & 20.8 & 17.8 & 13.8 & 6.0 & 5.4 & 13.0 & 70.6 \\
     &  & TSG 60 & 10.7 & 9.2 & 10.4 & 9.5 & 22.8 & 14.3 & 24.0 & 19.4 & 15.8 & 5.6 & 3.7 & 13.2 & 62.7 \\
    \cmidrule(lr){2-16}
     & \multirow{3}{*}{120} & --- & 11.9 & 12.1 & 10.5 & 10.9 & 24.9 & 13.9 & 27.4 & 18.4 & 17.2 & 6.2 & 4.3 & 14.3 & 62.4 \\
     &  & ATG 1.5 & 9.2 & 6.6 & 8.4 & 8.2 & 17.7 & 9.7 & 19.8 & 13.8 & 10.8 & 4.8 & 3.4 & 10.2 & 66.0 \\
     &  & ATG 2.5 & 6.7 & 6.1 & 6.3 & 5.4 & 14.4 & \underline{7.2} & 15.6 & \underline{10.5} & 9.3 & 3.5 & 2.3 & 7.9 & 70.5 \\
    \midrule
    \multirow{7}{*}{\shortstack[l]{Latent-\\Bandits}} & \multirow{4}{*}{60} & --- & 28.2 & 21.1 & 28.2 & 16.5 & 39.2 & 24.1 & 45.9 & 35.8 & 29.4 & 11.8 & 12.2 & 26.6 & 62.7 \\
     &  & ATG 1.5 & 19.9 & 13.4 & 18.5 & 11.9 & 28.4 & 16.8 & 36.3 & 24.4 & 20.5 & 7.2 & 7.0 & 18.6 & 67.2 \\
     &  & ATG 2.5 & 16.7 & 10.9 & 16.9 & 10.0 & 24.4 & 12.9 & 26.5 & 20.2 & 16.9 & 5.6 & 4.8 & 15.1 & 72.0 \\
     &  & TSG 60 & 20.0 & 13.1 & 18.7 & 11.1 & 30.0 & 16.7 & 32.6 & 24.4 & 20.3 & 6.6 & 6.6 & 18.2 & 62.7 \\
    \cmidrule(lr){2-16}
     & \multirow{3}{*}{120} & --- & 17.0 & 12.5 & 15.9 & 12.7 & 26.1 & 13.6 & 34.5 & 22.0 & 16.3 & 6.5 & 6.3 & 16.7 & 62.6 \\
     &  & ATG 1.5 & 11.5 & 7.2 & 11.1 & 5.2 & 17.9 & 7.5 & 23.6 & 13.3 & 10.6 & 3.3 & 4.3 & 10.5 & 67.0 \\
     &  & ATG 2.5 & 9.3 & 6.5 & 9.4 & 5.6 & 15.8 & \textcolor{red}{\textbf{6.5}} & 19.4 & 10.7 & 9.2 & \underline{2.7} & 2.6 & 8.9 & 71.9 \\
    \midrule
    \multirow{7}{*}{\shortstack[l]{Latent-\\Square}} & \multirow{4}{*}{60} & --- & 38.3 & 40.7 & 34.3 & 21.6 & 58.5 & 36.8 & 57.5 & 43.6 & 40.7 & 24.6 & 18.4 & 37.7 & 58.9 \\
     &  & ATG 1.5 & 28.4 & 30.7 & 27.3 & 17.0 & 48.2 & 29.2 & 46.6 & 35.9 & 33.9 & 17.8 & 14.6 & 30.0 & 62.2 \\
     &  & ATG 2.5 & 23.3 & 24.6 & 21.0 & 14.1 & 41.0 & 26.3 & 39.6 & 29.9 & 26.3 & 15.1 & 10.7 & 24.7 & 66.3 \\
     &  & TSG 60 & 19.5 & 20.6 & 20.9 & 12.1 & 37.4 & 22.9 & 37.8 & 28.9 & 23.7 & 12.3 & 10.5 & 22.4 & 59.5 \\
    \cmidrule(lr){2-16}
     & \multirow{3}{*}{120} & --- & 24.0 & 25.9 & 21.4 & 13.2 & 41.4 & 19.8 & 43.3 & 29.8 & 25.6 & 11.4 & 9.4 & 24.1 & 59.1 \\
     &  & ATG 1.5 & 16.6 & 17.7 & 15.5 & 8.3 & 32.2 & 15.6 & 36.9 & 23.1 & 21.0 & 8.8 & 7.1 & 18.4 & 62.5 \\
     &  & ATG 2.5 & 12.4 & 15.0 & 13.9 & 7.8 & 25.1 & 11.2 & 28.9 & 17.5 & 15.8 & 7.7 & 6.9 & 14.7 & 66.6 \\
    \midrule
    \multirow{7}{*}{\shortstack[l]{Latent-\\SimBA}} & \multirow{4}{*}{60} & --- & 83.1 & 76.3 & 74.4 & 62.3 & 92.8 & 87.3 & 90.4 & 89.1 & 87.1 & 71.0 & 62.1 & 79.6 & \textcolor{red}{\textbf{57.6}} \\
     &  & ATG 1.5 & 75.2 & 68.4 & 67.3 & 58.2 & 88.4 & 82.1 & 83.5 & 83.5 & 80.0 & 64.1 & 56.4 & 73.4 & 59.8 \\
     &  & ATG 2.5 & 66.2 & 61.1 & 59.8 & 47.6 & 82.7 & 76.0 & 75.8 & 76.1 & 72.6 & 55.8 & 51.0 & 65.9 & 63.0 \\
     &  & TSG 60 & 55.3 & 48.6 & 47.9 & 36.1 & 73.7 & 63.1 & 68.2 & 67.6 & 63.9 & 42.1 & 38.5 & 55.0 & \underline{58.2} \\
    \cmidrule(lr){2-16}
     & \multirow{3}{*}{120} & --- & 72.7 & 67.9 & 63.5 & 52.7 & 87.8 & 78.4 & 85.9 & 81.3 & 77.8 & 59.1 & 51.4 & 70.8 & \textcolor{red}{\textbf{57.6}} \\
     &  & ATG 1.5 & 65.5 & 58.7 & 57.4 & 45.2 & 81.9 & 70.5 & 78.3 & 74.3 & 68.1 & 51.1 & 45.0 & 63.3 & 59.7 \\
     &  & ATG 2.5 & 54.5 & 49.3 & 51.2 & 37.9 & 72.5 & 63.8 & 68.3 & 66.6 & 61.7 & 45.6 & 39.0 & 55.5 & 63.1 \\
    \bottomrule[1.5pt]
  \end{tabular}
  }
\end{table*}

\subsection{Experimental Setup}
\label{sec:setup}

\paragraph{Dataset and models}
Following DiffAttack~\cite{DBLP:journals/pami/ChenCCZZS25}, we use the ImageNet-Compatible Dataset~\cite{DBLP:journals/corr/abs-1804-00097} of 1000 ImageNet validation images and the same eleven victims, comprising five CNNs, four vision Transformers and two MLP-Mixers, plus four adversarially trained models in Section~\ref{sec:defense}.

\paragraph{Attack methods}
We denote the six adapted attacks, NES, ZOsignSGD, Square, SimBA, SignHunter and Bandits, by the prefix Latent and evaluate each on its own, with ATG at scales 1.5 and 2.5, and with TSG at $Q_c=60$. We compare with the transfer-based attacks PI-FGSM, S$^2$I-FGSM, NCF, DiffAttack, ACA~\cite{DBLP:conf/nips/ChenLWJDZ23}, APA with its SG and GC variants, and NatADiff~\cite{DBLP:conf/iclr/CollinsVFM26}, which synthesizes each sample from the class label instead of perturbing the clean image, so its examples have no clean reference. For transfer-based attacks, we average each victim’s accuracy over all other models used as surrogates. We further compare with the decision-based Latent-HSJA~\cite{DBLP:conf/eccv/NaJK22}, which observes only the predicted label, starts from an image of another class and moves along the decision boundary towards the original image. We adapt it to our framework by running it on the DDIM-inverted latent and generating each candidate with the remaining denoising steps. As it follows a different attack paradigm, it is not the main focus of our comparison, and we evaluate it separately on ResNet-50.

\paragraph{Implementation details}
We use the base model of Stable Diffusion v2.0~\cite{DBLP:conf/cvpr/RombachBLEO22} with 20 DDIM steps and $w=2.5$. Following DiffAttack, each image is inverted over the final $T=5$ steps, in which the attack is active, and the latent perturbation is bounded by $\varepsilon=0.1$ in the $L_\infty$ norm unless stated otherwise. We report the top-1 accuracy of the victim, where lower values indicate stronger attacks, and FID~\cite{DBLP:conf/nips/HeuselRUNH17} against the reference statistics of DiffAttack, averaged over the eleven victim-specific runs for our attacks.

\subsection{Performance of Latent Score-Based Attacks}
\label{sec:comparison}
As shown in Table~\ref{tab:main}, under both budgets, every unguided latent score-based attack achieves lower FID than all baselines, with FID between 57.6 and 62.7, whereas the baseline with the lowest FID, ACA, reaches 64.8. Several of our attacks also outperform the strongest transfer-based baseline using limited victim queries. With $Q=60$ queries per active step, the strongest strategy, Latent-ZOsignSGD, lowers the average accuracy to 21.3\%, below that of APA-SG, the second-strongest transfer-based attack, on all eleven victims. With $Q=120$, 4 of the 6 strategies, NES, ZOsignSGD, SignHunter and Bandits, outperform on average the strongest transfer-based attack APA-GC, which has the worst FID of 72.0 among all unrestricted attacks, with accuracies of 12.1\% to 16.7\% against 17.7\%. This advantage may arise from the scores of the victim, which steer the search towards its own decision boundary, and to the remaining denoising steps, which refine each latent perturbation into natural image content.

Strategies that perturb all latent coordinates in every query are the most effective. The gradient estimators NES and ZOsignSGD and the sign-based SignHunter reach average accuracies of 21.3\% to 24.2\% at $Q=60$ and 12.1\% to 14.3\% at $Q=120$, followed by Bandits. Strategies built on local or sparse updates fall behind, with Square at 37.7\% and 24.1\% and SimBA, which perturbs a single coordinate at a time, at 79.6\% and 70.8\%. These results favor dense latent updates under the tested query budgets.

\subsection{Performance of Adaptive-Target Guidance}
\label{sec:guidance}
Adaptive-target guidance consistently strengthens every strategy under both budgets in Table~\ref{tab:main}, lowering the average accuracy by 4.1\% to 8.0\% across all latent score-based attacks at $s=1.5$ and by 5.6\% to 15.3\% at $s=2.5$. With $s=2.5$, Latent-ZOsignSGD reaches 6.1\% at $Q=120$, about a third of the accuracy of APA-GC, and 9 of the 12 settings outperform APA-GC, compared with 4 without guidance.

This gain is accompanied by FID increases across all latent score-based attacks of 2.1 to 4.5 at $s=1.5$ and by 5.4 to 9.3 at $s=2.5$. At $s=1.5$, Latent-NES and Latent-ZOsignSGD already reach 8.5\% and 7.9\% at $Q=120$ with FID of 63.2 and 63.0, still below every transfer-based attack. ATG is therefore highly effective at boosting latent score-based attacks, with a trade-off between attack effectiveness and FID.

\subsection{Performance of Target-Search Guidance}
\label{sec:tsg}
Target-search guidance strengthens every strategy with small changes in FID. Compared with the unguided attacks at $Q=60$, it lowers the average accuracy by 5.9\% to 24.6\% yet raises FID by at most 0.7, whereas adaptive-target guidance raises FID by 2.1 to 9.3. Under the same latent budget, TSG attains lower accuracy than ATG at $s=1.5$ for five of the six strategies, with an FID that is 1.6 to 4.5 lower. With TSG, Latent-SignHunter, Latent-ZOsignSGD and Latent-NES reach 13.2\%, 15.4\% and 16.3\% at FID of 62.7, 60.1 and 60.1, outperforming APA-GC in both accuracy and image quality.

Since TSG requires up to 61 additional victim queries per step on the condition, we further compare it with the unguided attacks at $Q=120$, which use a comparable total query budget. The search-based strategies benefit more from devoting half of the budget to the condition, as SignHunter, Square and SimBA reach 13.2\%, 22.4\% and 55.0\% against 14.3\%, 24.1\% and 70.8\%, whereas the gradient estimators NES, ZOsignSGD and Bandits gain more from a larger latent budget. In all cases, FID stays within 0.7 of the unguided attacks. TSG therefore mitigates the quality degradation of ATG while still strengthening the attack, offering the best balance between effectiveness and stealthiness among our variants.

\subsection{Comparison with Decision-Based Attacks}
\label{sec:decision}
\begin{table}[t]
  \centering
  \caption{\label{tab:hsja}%
Comparison with decision-based Latent-HSJA on ResNet-50, with top-1 accuracy (\%, $\downarrow$) and FID ($\downarrow$). Queries are counted per image, i.e., $(T+1)Q$ with $Q=60$ and $Q=120$ for Latent-ZOsignSGD. As in the original work, Latent-HSJA places no bound on the latent perturbation. Best in \textcolor{red}{\textbf{red}}, second best \underline{underlined}.
  }
  \footnotesize
  \begin{tabular}{lccc}
    \toprule[1.5pt]
    Attack & Queries & Acc. & FID \\
    \midrule
    \multirow{5}{*}{Latent-HSJA} & 360 & 37.9 & 67.5 \\
     & 720 & 39.4 & 66.0 \\
     & 1440 & 41.8 & 64.4 \\
     & 3600 & 41.3 & 61.6 \\
     & 7200 & 45.2 & \underline{60.9} \\
    \midrule
    \multirow{2}{*}{Latent-ZOsignSGD} & 360 & \underline{20.8} & \textcolor{red}{\textbf{59.6}} \\
     & 720 & \textcolor{red}{\textbf{10.3}} & \textcolor{red}{\textbf{59.6}} \\
    \bottomrule[1.5pt]
  \end{tabular}
\end{table}
Under the same total query budget on ResNet-50, Latent-HSJA reaches 37.9\% at 360 queries with an FID of 67.5, as reported in Table~\ref{tab:hsja}. Its examples lie on the decision boundary of the victim, and many of them return to the original class once saved as 8-bit images, so the accuracy rises to 45.2\% as more queries move them closer to the original image, while FID decreases to 60.9. In contrast, Latent-ZOsignSGD reaches 20.8\% and 10.3\% with 360 and 720 queries at an FID of 59.6. In this comparison, Latent-ZOsignSGD achieves lower accuracy and FID than Latent-HSJA, even when the latter receives ten times the query budget.

\begin{figure*}[!t]
  \centering
  \includegraphics[width=\linewidth]{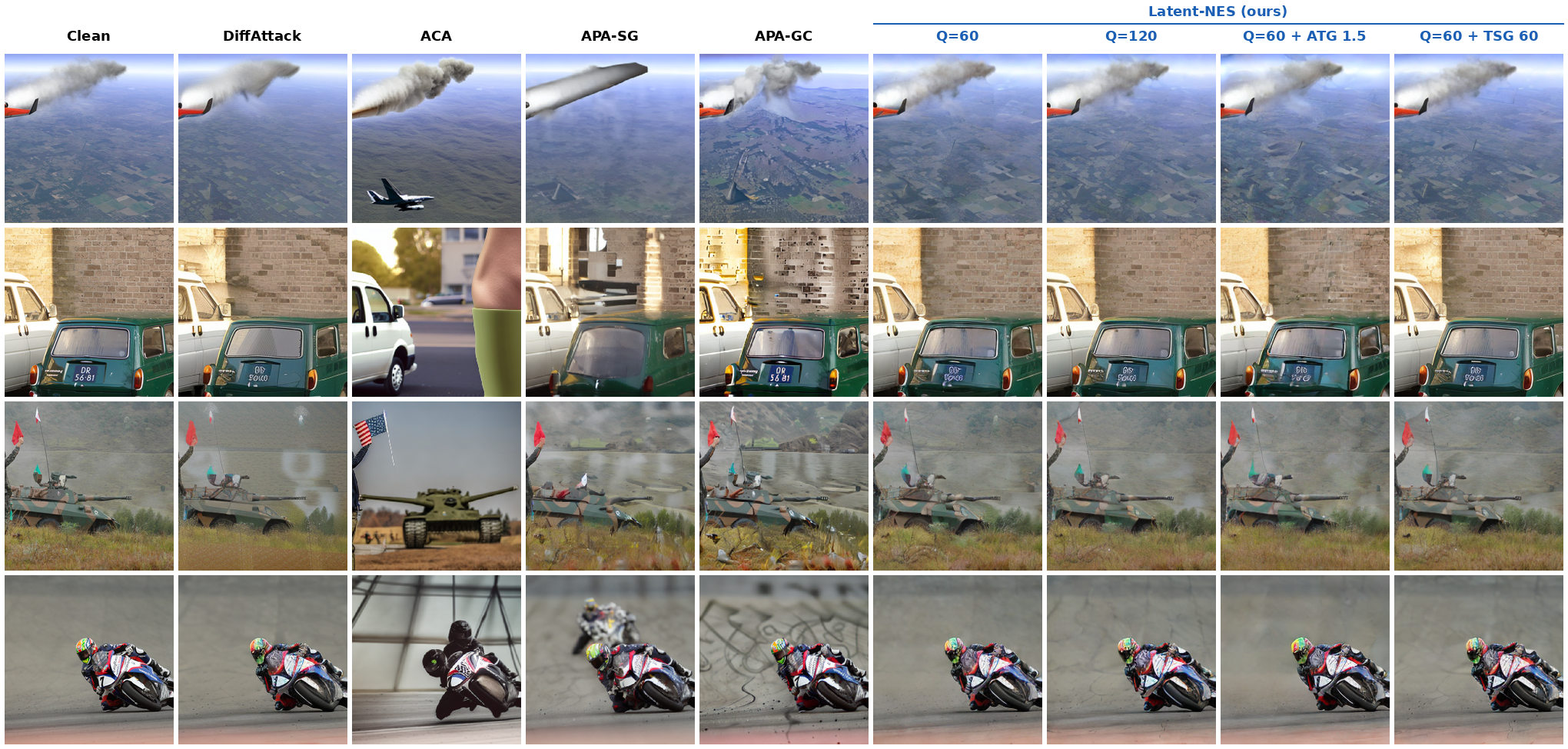}
  \caption{Qualitative comparison with ResNet-50 as the victim and as the white-box surrogate of the transfer-based baselines.}
  \label{fig:qualitative}
\end{figure*}

\subsection{Evaluation on Adversarially Trained Models}
\label{sec:defense}
\begin{table}[t]
  \centering
  \caption{\label{tab:at}%
    Top-1 accuracy (\%, $\downarrow$) of four adversarially trained models and FID ($\downarrow$). Transfer-based attacks are averaged over the nine surrogate models. Best in \textcolor{red}{\textbf{red}}, second best \underline{underlined}.
  }
  \setlength{\tabcolsep}{3pt}
  \resizebox{\columnwidth}{!}{
    \begin{tabular}{lcccccccc}
    \toprule[1.5pt]
    \multirow{2}{*}{Attack} & \multirow{2}{*}{$Q$} & \multirow{2}{*}{Guidance} & Inc- & Inc- & Inc- & IncRes- & \multirow{2}{*}{Avg.} & \multirow{2}{*}{FID} \\
    & & & v3$_{\mathrm{adv}}$ & v3$_{\mathrm{ens3}}$ & v3$_{\mathrm{ens4}}$ & v2$_{\mathrm{ens}}$ & & \\
    \midrule
    Clean & --- & --- & 84.4 & 79.4 & 78.7 & 90.3 & 83.2 & 57.8 \\
    \midrule
    DiffAttack & --- & --- & 42.6 & 40.9 & 40.9 & 51.2 & 43.9 & 65.4 \\
    ACA & --- & --- & 35.4 & 32.6 & 33.2 & 40.4 & 35.4 & 64.8 \\
    APA-SG & --- & --- & 24.5 & 24.1 & 24.2 & 31.3 & 26.0 & 65.2 \\
    APA-GC & --- & --- & 14.9 & 16.1 & 15.5 & 19.6 & 16.5 & 72.0 \\
    NatADiff & --- & --- & 33.6 & 34.4 & 33.4 & 36.7 & 34.5 & 66.1 \\
    \midrule
    \multirow{7}{*}{\shortstack[l]{Latent-\\ZOsignSGD}} & \multirow{4}{*}{60} & --- & 15.9 & 12.5 & 12.1 & 21.3 & 15.5 & \textcolor{red}{\textbf{59.2}} \\
     &  & ATG 1.5 & 10.6 & 8.9 & 8.7 & 16.2 & 11.1 & 63.1 \\
     &  & ATG 2.5 & 8.7 & 8.4 & 6.6 & 12.4 & 9.0 & 67.9 \\
     &  & TSG 60 & 9.9 & 8.8 & 8.0 & 14.7 & 10.4 & \underline{59.7} \\
    \cmidrule(lr){2-9}
     & \multirow{3}{*}{120} & --- & 10.1 & 8.1 & 7.0 & 13.2 & 9.6 & \textcolor{red}{\textbf{59.2}} \\
     &  & ATG 1.5 & \textcolor{red}{\textbf{5.7}} & \underline{4.3} & \underline{4.8} & \underline{9.4} & \underline{6.1} & 63.0 \\
     &  & ATG 2.5 & \underline{6.2} & \textcolor{red}{\textbf{4.1}} & \textcolor{red}{\textbf{4.0}} & \textcolor{red}{\textbf{7.5}} & \textcolor{red}{\textbf{5.5}} & 67.8 \\
    \bottomrule[1.5pt]
  \end{tabular}
  }
\end{table}
On the four adversarially trained models: DiffAttack and ACA leave average accuracies of 43.9\% and 35.4\% on the four defended models of Table~\ref{tab:at}, and even APA-GC leaves 16.5\%. Since adversarial training is a property of the victim itself, our attack simply queries the defended model, and Latent-ZOsignSGD without guidance already reaches 15.5\% at $Q=60$ and 9.6\% at $Q=120$ with an FID of 59.2. Guidance further lowers the accuracy to 5.5\% with ATG at $s=2.5$ and $Q=120$, while TSG reaches 10.4\% at $Q=60$ with an FID of 59.7. These results show that the latent score-based attack remains effective on the four evaluated adversarially trained models. 

\subsection{Ablation Study}
\label{sec:ablation}
\begin{table}[t]
  \centering
  \caption{\label{tab:ablation}%
    Ablation of the components of Latent-ZOsignSGD at $Q=60$, with ATG at $s=1.5$ and TSG at $Q_c=60$. We report the top-1 accuracy (\%, $\downarrow$) of ResNet-50 and ViT-B and the FID ($\downarrow$) averaged over both. Latent denotes the score-based latent perturbation. Best in \textcolor{red}{\textbf{red}}, second best \underline{underlined}. Tables 1 and 4 report separate runs, values for identical configurations may therefore differ.
  }
  \footnotesize
  \begin{tabular}{cccccc}
    \toprule[1.5pt]
    Latent & ATG & TSG & ResNet-50 & ViT-B & FID \\
    \midrule
    \checkmark &  &  & 20.8 & 24.0 & 59.9 \\
     & \checkmark &  & 85.2 & 88.8 & \underline{59.5} \\
     &  & \checkmark & 67.9 & 76.7 & \textcolor{red}{\textbf{58.3}} \\
    \checkmark & \checkmark &  & \underline{14.2} & \textcolor{red}{\textbf{17.3}} & 63.2 \\
    \checkmark &  & \checkmark & \textcolor{red}{\textbf{13.9}} & \underline{18.7} & 60.5 \\
    \bottomrule[1.5pt]
  \end{tabular}
\end{table}
Table~\ref{tab:ablation} isolates the latent perturbation, ATG and TSG of Latent-ZOsignSGD at $Q=60$. The latent perturbation alone lowers the accuracy of ResNet-50 and ViT-B to 20.8\% and 24.0\%, whereas either guidance alone leaves between 67.9\% and 88.8\%. Combined with the latent perturbation, ATG lowers the accuracy by a further 6.6\% and 6.7\%, and TSG by 6.9\% and 5.3\%, while TSG raises the FID by only 0.6 and ATG by 3.3. The latent perturbation therefore provides most of the attack strength, while both guidance strategies are most effective in combination with it.

\begin{figure}[t]
  \centering
  \includegraphics[width=0.9\linewidth]{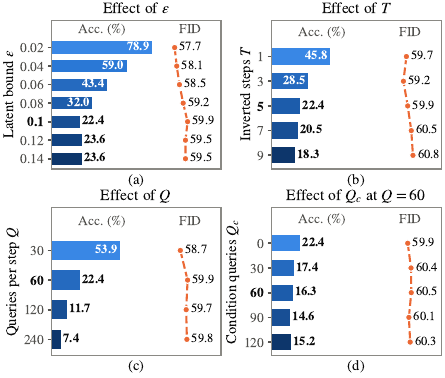}
  \caption{Parameter analysis of Latent-ZOsignSGD without guidance, varying one setting at a time from the default $\varepsilon=0.1$, $T=5$ and $Q=60$, shown in bold. In (d), TSG adds $Q_c$ condition queries per active step to $Q=60$. Bars show the top-1 accuracy (\%, $\downarrow$) and dots the FID ($\downarrow$), both averaged over ResNet-50 and ViT-B.}
  \label{fig:params}\label{fig:eps_ablation}
\end{figure}

\subsection{Parameter Analysis}
\label{sec:params}
Increasing the latent bound in Figure~\ref{fig:params}(a) from $\varepsilon=0.02$ to $0.1$ lowers the accuracy from 78.9\% to 22.4\%, while the FID rises only from 57.7 to 59.9, and larger bounds bring no further gain. More inverted steps and more queries per step both strengthen the attack at a proportional query cost, lowering the accuracy from 45.8\% at $T=1$ to 18.3\% at $T=9$ and from 53.9\% at $Q=30$ to 7.4\% at $Q=240$. With $Q=60$ fixed, TSG lowers the accuracy from 22.4\% to 16.3\% at $Q_c=60$ and to 14.6\% at $Q_c=90$, with no further gain at $Q_c=120$, while raising the FID by at most 0.6. Across all settings, the FID stays between 57.7 and 60.8. Across the tested ranges, accuracy varies substantially while FID changes modestly.

\subsection{Qualitative Results}
Among the examples in Figure~\ref{fig:qualitative}, ACA regenerates the image content, inserting an additional aircraft and re-rendering the scenes around the van, the tank and the motorcyclist, while APA-SG and APA-GC keep the layout but add visible artifacts such as a distorted brick wall and scribble-like patterns on the racetrack. DiffAttack stays close to the clean images at the cost of weaker attacks. Our examples largely preserve the object identity, pose and background, with changes mainly affecting fine textures such as the license plate and the terrain, and ATG introduces slightly stronger texture changes than a larger budget or TSG, consistent with their FID.

%% file: sec/6_conclusion.tex
\section{Conclusion}
In this paper, we adapted score-based black-box attacks to the reverse diffusion process to craft unrestricted adversarial examples tailored to each victim model, and strengthened them with adaptive-target guidance (ATG) and target-search guidance (TSG). Our attacks surpassed the strongest transfer-based attacks in both effectiveness and image quality, also on adversarially trained models. ATG yielded the strongest attacks, while TSG improved effectiveness at nearly unchanged image quality.

%% file: sec/7_llm_statement.tex
\section*{LLM Usage Statement}
We used Claude to assist with English grammar correction, language polishing, and generate visualization code. Its use was limited to improving the
wording of the manuscript and visualizing experimental results.

%% file: sec/X_suppl.tex
\clearpage
\setcounter{page}{1}
\maketitlesupplementary

\renewcommand{\dbltopfraction}{0.92}\renewcommand{\topfraction}{0.9}\renewcommand{\textfraction}{0.07}
\renewcommand{\floatpagefraction}{0.8}\renewcommand{\dblfloatpagefraction}{0.8}
\setcounter{dbltopnumber}{3}\setcounter{topnumber}{3}

In this supplementary material, we provide an extended study of our method. We first report the performance of combining ATG and TSG in Sec. \ref{sec:supp-combo}, and investigate our alternating strategy to generate adversarial examples during the reverse diffusion process in Sec. \ref{sec:supp-analysis}. We compare the computational efficiency of existing methods to ours in Sec. \ref{sec:supp-time}, and assess the robustness of our method to random seeds in Sec. \ref{sec:supp-seed}. We further extend image quality comparisons from FID to more image quality metrics in Sec. \ref{sec:supp-iqa}, followed by implementation details of adapting different score-based attacks to latent space in Sec. \ref{sec:supp-impl}. We offer a simplified example of our algorithm for better visualization in Sec. \ref{sec:toy} and qualitative results in Sec. \ref{sec:supp-qual}.

\section{Interaction Between ATG and TSG}
\label{sec:supp-combo}
We examine whether ATG and TSG can be used together. Applying both at full strength lowers the accuracy of the victim to 9.6\% on ResNet-50 and 14.1\% on ViT-B, 3.2\% below the stronger of the two guidance methods on each victim, with 670 queries per image, five more than TSG. This gain comes with a higher FID and LPIPS than either guidance method alone.

Each reverse step applies the deterministic DDIM update
\begin{equation}
\begin{split}
  x_{t-1}={}&\sqrt{\bar\alpha_{t-1}}\,\hat{x}_{0|t}+\sqrt{1-\bar\alpha_{t-1}}\,\tilde{\boldsymbol{\epsilon}}_\theta(x_t),\\
  \hat{x}_{0|t}={}&\bigl(x_t-\sqrt{1-\bar\alpha_t}\,\tilde{\boldsymbol{\epsilon}}_\theta(x_t)\bigr)/\sqrt{\bar\alpha_t},
\end{split}
\end{equation}
with the guided noise prediction $\tilde{\boldsymbol{\epsilon}}_\theta$. Since $\boldsymbol{\epsilon}_\theta(x_t,t,\mathbf{c})\approx-\sqrt{1-\bar\alpha_t}\,\nabla_{x_t}\log p(x_t\mid\mathbf{c})$, the difference $\boldsymbol{\epsilon}_\theta(x_t,t,\mathbf{c})-\boldsymbol{\epsilon}_\theta(x_t,t,\varnothing)$ is proportional to the negative gradient of the implicit classifier $\log p(\mathbf{c}\mid x_t)$, so adding it to the noise prediction with a positive weight steers the sample towards the condition $\mathbf{c}$. Classifier-free guidance adds this difference for the original condition $\mathbf{c}_{\mathrm{ori}}$ with weight $w$. TSG replaces $\mathbf{c}_{\mathrm{ori}}$ with the condition $\mathbf{c}_{\mathrm{adv}}$ searched with $Q_c$ queries, and ATG adds the difference for the condition $\mathbf{c}_{\mathrm{tar}}$ of the highest-probability class other than $y_{\mathrm{ori}}$, with weight $s$. Both enter one noise prediction,
\begin{equation}
\begin{split}
  \tilde{\boldsymbol{\epsilon}}_\theta(x_t)=\ &\boldsymbol{\epsilon}_\theta(x_t,t,\varnothing)+w\bigl[a\,\boldsymbol{\epsilon}_\theta(x_t,t,\mathbf{c}_{\mathrm{adv}})\\
  &+(1-a)\,\boldsymbol{\epsilon}_\theta(x_t,t,\mathbf{c}_{\mathrm{ori}})-\boldsymbol{\epsilon}_\theta(x_t,t,\varnothing)\bigr]\\
  &+b\,s\bigl[\boldsymbol{\epsilon}_\theta(x_t,t,\mathbf{c}_{\mathrm{tar}})-\boldsymbol{\epsilon}_\theta(x_t,t,\varnothing)\bigr],
\end{split}
  \label{eq:supp-combo}
\end{equation}
where $a=1,b=0$ recovers TSG and $a=0,b=1$ recovers ATG. Tab.~\ref{tab:supp-combo} interpolates between them with $a=1-\lambda$ and $b=\lambda$, and applies both in full with $a=b=1$ as well, for Latent-ZOsignSGD with $Q=60$, $Q_c=60$ and $s=1.5$. The interpolation raises the FID steadily from TSG towards ATG, from 59.9 to 63.6 on ResNet-50 and from 60.7 to 63.5 on ViT-B, whereas the accuracy shows no consistent trend, ranging from 12.6\% to 14.6\% on ResNet-50 and from 17.1\% to 18.7\% on ViT-B. Only the full combination lowers the accuracy below both guidance methods on both victims by margin, at an FID 0.4 and 1.0 above ATG. Applying both in full therefore gives the lowest accuracy observed, at a moderate cost in fidelity.

\begin{table}[!htb]
  \centering
  \small
  \caption{\textbf{Top-1 accuracy} (\%, $\downarrow$) of the victim, \textbf{FID} ($\downarrow$), \textbf{LPIPS} ($\downarrow$), \textbf{SSIM} ($\uparrow$) and \textbf{number of victim queries} per image when ATG and TSG are combined by Eq.~\eqref{eq:supp-combo}, for Latent-ZOsignSGD with $Q=60$, $Q_c=60$ and $s=1.5$.}
  \label{tab:supp-combo}
  \resizebox{\ifdim\width>\linewidth\linewidth\else\width\fi}{!}{
    \begin{tabular}{llccccc}
    \toprule
    Victim & Guidance & Queries & Victim acc. & FID & LPIPS & SSIM \\
    \midrule
    \multirow{6}{*}{Res50} & TSG ($\lambda=0$) & 665 & 12.8 & 59.9 & 0.112 & 0.683 \\
     & $\lambda=0.25$ & 670 & 14.6 & 60.3 & 0.113 & 0.682 \\
     & $\lambda=0.5$ & 670 & 13.8 & 61.2 & 0.117 & 0.678 \\
     & $\lambda=0.75$ & 670 & 12.6 & 62.0 & 0.124 & 0.672 \\
     & ATG ($\lambda=1$) & 365 & 13.9 & 63.6 & 0.133 & 0.664 \\
     & ATG + TSG & 670 & 9.6 & 64.0 & 0.137 & 0.663 \\
    \midrule
    \multirow{6}{*}{ViT-B} & TSG ($\lambda=0$) & 665 & 18.7 & 60.7 & 0.117 & 0.678 \\
     & $\lambda=0.25$ & 670 & 18.6 & 60.9 & 0.118 & 0.677 \\
     & $\lambda=0.5$ & 670 & 17.5 & 61.9 & 0.122 & 0.673 \\
     & $\lambda=0.75$ & 670 & 17.1 & 62.3 & 0.130 & 0.666 \\
     & ATG ($\lambda=1$) & 365 & 17.3 & 63.5 & 0.140 & 0.658 \\
     & ATG + TSG & 670 & 14.1 & 64.5 & 0.142 & 0.657 \\
    \bottomrule
    \end{tabular}
  }
\end{table}

\section{Alternating versus Fixed-Step Updates}
\label{sec:supp-analysis}
Our attack alternates guided denoising steps with score-based latent updates. We compare it with a fixed-step variant that spends the whole budget of $(T+1)Q$ queries on $x_T$ within the same $L_\infty$ bound and scores every query on the image obtained by running the remaining guided DDIM steps. With TSG, the variant searches the condition once with $TQ_c$ queries on the same rollouts.

Tab.~\ref{tab:supp-match} pairs fixed-step settings with alternating settings of similar accuracy on ResNet-50. With $Q=60$, the alternating attack makes two sign updates per active step, so its step size is set to $\varepsilon/2$ to reach the bound within them. The alternating attack reaches the accuracy of the fixed-step variant within 1.6\%, at an FID 0.3 lower and 0.9 higher, while it takes 4.9$\times$ and 4.5$\times$ less time on the same GPU. Alternating therefore attains the accuracy and image quality of the fixed-step variant at a fraction of its cost.

\begin{table*}[t]
  \centering
  \small
  \caption{\textbf{Top-1 accuracy} (\%, $\downarrow$) of ResNet-50, \textbf{FID} ($\downarrow$) and \textbf{attack time per image} (s, $\downarrow$) of the fixed-step variant and of the alternating setting with similar accuracy, with the \textbf{speed-up} of alternating, for Latent-ZOsignSGD with $Q=60$ and no guidance. Times are measured on an NVIDIA L40 GPU.}
  \label{tab:supp-match}
  \begin{tabular}{cccccccccc}
    \toprule
    \multicolumn{4}{c}{Fixed-step} & \multicolumn{5}{c}{Alternating} & \multirow{2}{*}{Speed-up} \\
    \cmidrule(lr){1-4} \cmidrule(lr){5-9}
    $\varepsilon$ & Res50 & FID & Time & $\varepsilon$ & Step & Res50 & FID & Time & \\
    \midrule
    0.1 & 34.5 & 58.6 & 56.4 & 0.07 & 0.035 & 36.1 & 58.3 & 11.6 & 4.9$\times$ \\
    0.15 & 20.9 & 58.9 & 46.5 & 0.1 & 0.05 & 21.6 & 59.8 & 10.3 & 4.5$\times$ \\
    \bottomrule
  \end{tabular}
\end{table*}

At the same query budget, alternating lowers the accuracy of the victim by 5.5\% to 19.5\% in Tab.~\ref{tab:supp-fsr} and takes 4.7$\times$ to 9.8$\times$ less time per image, measured on the same images and GPU type. The fixed-step variant attains an FID that is up to 2.1 lower, since its perturbation is bounded once rather than at every step. These results suggest that ATG and TSG can provide complementary guidance.

\begin{table*}[t]
  \centering
  \small
  \caption{\textbf{Top-1 accuracy} (\%, $\downarrow$) of the victim, \textbf{FID} ($\downarrow$), \textbf{LPIPS} ($\downarrow$), \textbf{SSIM} ($\uparrow$) and \textbf{attack time per image} (s, $\downarrow$) of the alternating and fixed-step variants of Latent-ZOsignSGD with $Q=60$ and $\varepsilon=0.1$. Alternating is averaged over three seeds. Times are measured on a Blackwell GPU for ResNet-50 and an L40 GPU for ViT-B.}
  \label{tab:supp-fsr}
    \begin{tabular}{lllccccc}
    \toprule[1.5pt]
    Victim & Guidance & Variant & Victim acc. & FID & LPIPS & SSIM & Time \\
    \midrule
    Res50 & --- & Alternating & 21.8 & 59.7 & 0.108 & 0.683 & 6.4 \\
     & & Fixed-step & 34.5 & 58.6 & 0.096 & 0.702 & 39.8 \\
    \cmidrule(lr){2-8}
     & ATG 1.5 & Alternating & 13.7 & 62.9 & 0.133 & 0.664 & 6.3 \\
     & & Fixed-step & 25.9 & 61.2 & 0.123 & 0.679 & 61.7 \\
    \cmidrule(lr){2-8}
     & TSG 60 & Alternating & 14.5 & 60.3 & 0.111 & 0.683 & 13.8 \\
     & & Fixed-step & 20.0 & 60.4 & 0.116 & 0.686 & 64.5 \\
    \midrule
    ViT-B & --- & Alternating & 24.2 & 60.1 & 0.114 & 0.678 & 11.2 \\
     & & Fixed-step & 43.7 & 58.0 & 0.098 & 0.701 & 63.5 \\
    \cmidrule(lr){2-8}
     & ATG 1.5 & Alternating & 17.1 & 63.5 & 0.139 & 0.658 & 11.0 \\
     & & Fixed-step & 35.6 & 61.7 & 0.125 & 0.677 & 99.3 \\
    \cmidrule(lr){2-8}
     & TSG 60 & Alternating & 18.3 & 60.5 & 0.116 & 0.678 & 21.9 \\
     & & Fixed-step & 31.8 & 59.9 & 0.117 & 0.685 & 103.2 \\
    \bottomrule[1.5pt]
    \end{tabular}
\end{table*}

\section{Computational Efficiency}
\label{sec:supp-time}

We compare per-image time cost across all methods. Tab.~\ref{tab:supp-time} lists the victim queries and the attack time per image, excluding model loading. All attacks are timed with ResNet-50 and ViT-B/16 as the victim or surrogate on an NVIDIA L40 GPU and with ResNet-50 on an NVIDIA RTX Pro 6000 Blackwell GPU. The time of ACA includes its per-image null-text inversion and that of APA its per-image visual alignment, since both precede the attack. Without guidance, Latent-ZOsignSGD is 1.7$\times$ to 1.8$\times$ faster than DiffAttack in the three settings and 4.5$\times$ to 11.1$\times$ faster than Latent-HSJA with the same 360 queries, APA, NatADiff and ACA. Only PI-FGSM and S$^2$I-FGSM are faster, at an FID of 94.0 and 77.2 in Tab.~\ref{tab:main}. ATG adds one query and one noise prediction per active step and leaves the time almost unchanged, whereas TSG spends $Q_c$ additional queries and noise predictions per active step and increases the time by a factor of 2.0 to 2.2.

\begin{table}[!htb]
  \centering
  \small
  \caption{\textbf{Number of victim queries} and \textbf{attack time per image} (s, $\downarrow$), excluding model loading. Transfer-based attacks use the model of each column as the surrogate. The Blackwell and ViT-B times of Latent-ZOsignSGD are averaged over three seeds.}
  \label{tab:supp-time}
  \resizebox{\linewidth}{!}{
    \begin{tabular}{lccccc}
    \toprule
    \multirow{2}{*}{Attack} & \multirow{2}{*}{Guidance} & \multirow{2}{*}{Queries} & \multicolumn{2}{c}{L40} & Blackwell \\
    \cmidrule(lr){4-5} \cmidrule(lr){6-6}
     & & & Res50 & ViT-B & Res50 \\
    \midrule
    PI-FGSM & --- & --- & 0.09 & 0.10 & 0.04 \\
    S$^2$I-FGSM & --- & --- & 1.2 & 1.9 & 0.8 \\
    DiffAttack & --- & --- & 18.3 & 19.0 & 10.9 \\
    ACA & --- & --- & 115.3 & 117.2 & 63.2 \\
    APA-SG & --- & --- & 87.3 & 89.0 & 47.8 \\
    APA-GC & --- & --- & 82.9 & 83.6 & 46.6 \\
    NatADiff & --- & --- & 90.3 & 79.9 & 41.4 \\
    \midrule
    \multirow{2}{*}{Latent-HSJA} & --- & 360 & 50.8 & 50.7 & 33.6 \\
     & --- & 720 & 95.1 & 94.9 & 64.0 \\
    \midrule
    \multirow{4}{*}{\shortstack[l]{Latent-\\ZOsignSGD}} & --- & 360 & 10.3 & 11.2 & 6.5 \\
     & ATG 1.5 & 365 & 10.3 & 11.0 & 6.3 \\
     & TSG 60 & 665 & 22.4 & 21.9 & 13.9 \\
     & --- & 720 & 12.2 & 13.3 & 7.6 \\
    \bottomrule
    \end{tabular}
  }
\end{table}

\section{Robustness Across Seeds}
\label{sec:supp-seed}
\begin{table*}[t]
  \centering
  \small
  \caption{\textbf{Top-1 accuracy} (\%, $\downarrow$) of the eleven victims, its average and \textbf{FID} ($\downarrow$) of Latent-ZOsignSGD, as mean $\pm$ standard deviation over three seeds next to the results of Tab.~\ref{tab:main}.}
  \label{tab:supp-seed}
  \begin{tabular}{lcccccccc}
    \toprule
    & \multicolumn{2}{c}{$Q=60$} & \multicolumn{2}{c}{$Q=60$, ATG 1.5} & \multicolumn{2}{c}{$Q=60$, TSG 60} & \multicolumn{2}{c}{$Q=120$} \\
    \cmidrule(lr){2-3} \cmidrule(lr){4-5} \cmidrule(lr){6-7} \cmidrule(lr){8-9}
    Victim & Seeds & Tab.~\ref{tab:main} & Seeds & Tab.~\ref{tab:main} & Seeds & Tab.~\ref{tab:main} & Seeds & Tab.~\ref{tab:main} \\
    \midrule
    Res50 & 21.8$\pm$0.6 & 21.1 & 13.7$\pm$0.9 & 14.4 & 14.5$\pm$0.6 & 13.5 & 10.9$\pm$0.6 & 10.3 \\
    VGG19 & 16.4$\pm$0.9 & 16.8 & 11.0$\pm$0.9 & 11.0 & 11.5$\pm$0.4 & 11.4 & 8.5$\pm$0.5 & 9.2 \\
    MobV2 & 18.1$\pm$1.1 & 19.3 & 12.7$\pm$0.5 & 11.9 & 12.2$\pm$0.6 & 11.4 & 9.5$\pm$0.3 & 9.7 \\
    IncV3 & 10.8$\pm$0.3 & 10.8 & 7.5$\pm$0.8 & 7.0 & 7.1$\pm$0.3 & 6.7 & 5.0$\pm$0.4 & 5.3 \\
    ConvNeXt & 35.7$\pm$0.9 & 34.3 & 26.6$\pm$0.6 & 25.2 & 26.6$\pm$0.6 & 27.1 & 22.3$\pm$0.4 & 22.4 \\
    ViT-B & 24.2$\pm$0.8 & 24.7 & 17.1$\pm$0.8 & 15.6 & 18.3$\pm$1.0 & 19.0 & 13.4$\pm$1.2 & 13.1 \\
    Swin-B & 33.8$\pm$1.0 & 33.2 & 23.8$\pm$0.5 & 25.2 & 26.4$\pm$0.2 & 25.5 & 22.0$\pm$0.2 & 22.4 \\
    DeiT-B & 30.1$\pm$0.6 & 28.4 & 21.1$\pm$0.3 & 21.3 & 23.4$\pm$0.4 & 22.6 & 18.5$\pm$0.3 & 18.9 \\
    DeiT-S & 24.6$\pm$0.5 & 25.6 & 17.5$\pm$0.5 & 16.7 & 19.0$\pm$0.3 & 18.8 & 14.1$\pm$0.8 & 13.2 \\
    Mixer-B & 10.0$\pm$0.3 & 10.6 & 7.0$\pm$0.5 & 6.6 & 7.2$\pm$0.6 & 6.7 & 4.4$\pm$0.2 & 4.6 \\
    Mixer-L & 9.9$\pm$0.6 & 9.0 & 6.0$\pm$0.6 & 6.2 & 6.1$\pm$0.3 & 6.2 & 4.5$\pm$0.2 & 4.5 \\
    \midrule
    Avg. & 21.4$\pm$0.2 & 21.3 & 14.9$\pm$0.1 & 14.6 & 15.7$\pm$0.1 & 15.4 & 12.1$\pm$0.3 & 12.1 \\
    FID & 59.5$\pm$0.1 & 59.4 & 63.0$\pm$0.0 & 62.8 & 60.1$\pm$0.1 & 60.1 & 59.6$\pm$0.1 & 59.6 \\
    \bottomrule
  \end{tabular}
\end{table*}

To validate our results under different seeds, we repeat Latent-ZOsignSGD for the seeds 0, 1 and 2 with $Q=60$ without guidance, with ATG at $s=1.5$ and with TSG at $Q_c=60$, and with $Q=120$ without guidance, on all eleven victims. Tab.~\ref{tab:supp-seed} reports the mean and standard deviation over the seeds next to the results of Tab.~\ref{tab:main}. The standard deviation of the accuracy is at most 1.2\% for a single victim and 0.3\% for the average, and that of FID is at most 0.1. The results of Tab.~\ref{tab:main} lie within 0.3\% of the mean in average accuracy and within 0.2 in FID, so the differences between settings exceed the variation across seeds.

\section{Extended Image Quality Evaluation}
\label{sec:supp-iqa}
We compare image quality of results from different methods. Tab.~\ref{tab:supp-iqa} complements FID, which compares distributions, with metrics computed on every image. The full-reference metrics compare each adversarial example with its clean image. PSNR measures the pixel error, SSIM, MS-SSIM and FSIM the similarity of local structures, and LPIPS, DISTS and TOPIQ-FR the similarity of deep features, which follows human judgments of perceptual similarity. The no-reference metrics NIQE and BRISQUE measure how far a single image deviates from the statistics of natural images. NatADiff differs from the other attacks in how it generates examples, as it synthesizes each example from the class label instead of editing the clean image, so its full-reference metrics compare two different images of the same class. All metrics are computed at $224\times224$. As for FID in Tab.~\ref{tab:main}, each transfer-based attack is averaged over the examples crafted on its nine surrogates, and each of our attacks over the examples generated against the eleven victims.

Without guidance, Latent-NES, Latent-ZOsignSGD, Latent-Square and Latent-SimBA outperform every transfer-based attack on all seven full-reference metrics, and Latent-SimBA attains the best value on each of them. Every latent attack without guidance attains a DISTS of 0.085 to 0.148, below the 0.155 to 0.410 of the transfer-based attacks. ATG lowers the fidelity as its scale grows, increasing LPIPS by 0.019 to 0.028 at $s=1.5$ and by 0.043 to 0.060 at $s=2.5$, whereas TSG changes LPIPS by at most 0.012. Doubling the budget to $Q=120$ changes PSNR by at most 0.09 and LPIPS by at most 0.003. On the no-reference metrics, all our settings attain a NIQE of 4.90 to 5.74 and a BRISQUE of 14.90 to 19.00, close to the 5.64 and 17.58 of the clean images, whereas the transfer-based attacks reach up to 7.52 and 26.90. Our attacks therefore preserve both the content of the clean image and the statistics of natural images, and ATG trades part of this fidelity for a stronger attack.

\begin{table*}[p]
  \centering
  \caption{\textbf{Full-reference and no-reference image quality metrics} of the adversarial examples on the ImageNet-Compatible dataset. Full-reference metrics compare each adversarial example with its clean image, and NIQE and BRISQUE are no-reference. Transfer-based attacks are averaged over their nine surrogates, and Latent-HSJA and our attacks over the eleven victims. Latent-HSJA uses 360 or 720 queries per image, as our attacks with $Q=60$ and $Q=120$, and places no bound on the latent perturbation. $^\dagger$NatADiff synthesizes each example from the class label instead of editing the clean image, so its full-reference metrics compare it with a different image of the same class. Best in \textcolor{red}{\textbf{red}}, second best \underline{underlined}.}
  \label{tab:supp-iqa}
  \footnotesize
  \resizebox{0.96\textwidth}{!}{
    \begin{tabular}{lccccccccccc}
    \toprule[1.5pt]
    \multirow{2}{*}{Attack} & \multirow{2}{*}{$Q$} & \multirow{2}{*}{Guidance} & \multicolumn{7}{c}{Full-reference} & \multicolumn{2}{c}{No-reference} \\
    \cmidrule(lr){4-10} \cmidrule(lr){11-12}
    & & & PSNR$\uparrow$ & SSIM$\uparrow$ & MS-SSIM$\uparrow$ & FSIM$\uparrow$ & LPIPS$\downarrow$ & DISTS$\downarrow$ & TOPIQ-FR$\uparrow$ & NIQE$\downarrow$ & BRISQUE$\downarrow$ \\
    \midrule
    Clean & --- & --- & --- & --- & --- & --- & --- & --- & --- & 5.64 & 17.58 \\
    \midrule
    DiffAttack & --- & --- & 22.51 & 0.664 & 0.888 & 0.835 & 0.143 & 0.155 & 0.613 & 5.83 & 18.00 \\
    ACA & --- & --- & 17.96 & 0.565 & 0.727 & 0.726 & 0.386 & 0.259 & 0.352 & 7.52 & 26.47 \\
    APA-SG & --- & --- & 19.48 & 0.650 & 0.805 & 0.785 & 0.250 & 0.197 & 0.433 & 6.16 & 26.90 \\
    APA-GC & --- & --- & 21.09 & 0.667 & 0.853 & 0.813 & 0.220 & 0.203 & 0.481 & 5.50 & 24.40 \\
    NatADiff$^\dagger$ & --- & --- & 9.06 & 0.179 & 0.120 & 0.519 & 0.738 & 0.410 & 0.240 & 6.74 & 18.06 \\
    \midrule
    \multirow{2}{*}{Latent-HSJA} & 360 & --- & 21.16 & 0.610 & 0.833 & 0.801 & 0.212 & 0.197 & 0.547 & 8.07 & 22.68 \\
     & 720 & --- & 21.76 & 0.627 & 0.853 & 0.814 & 0.184 & 0.177 & 0.576 & 7.13 & 20.53 \\
    \midrule
    \multirow{7}{*}{\shortstack[l]{Latent-\\NES}} & \multirow{4}{*}{60} & --- & 23.20 & 0.682 & 0.902 & 0.850 & 0.109 & 0.119 & 0.717 & 5.19 & 16.98 \\
     &  & ATG 1.5 & 22.39 & 0.663 & 0.884 & 0.837 & 0.134 & 0.144 & 0.653 & 5.17 & 17.25 \\
     &  & ATG 2.5 & 21.56 & 0.638 & 0.861 & 0.822 & 0.165 & 0.168 & 0.583 & 5.16 & 17.88 \\
     &  & TSG 60 & 23.21 & 0.682 & 0.902 & 0.849 & 0.112 & 0.123 & 0.704 & 5.28 & 16.97 \\
    \cmidrule(lr){2-12}
     & \multirow{3}{*}{120} & --- & 23.24 & 0.683 & 0.903 & 0.850 & 0.108 & 0.118 & 0.718 & 5.19 & 16.96 \\
     &  & ATG 1.5 & 22.41 & 0.663 & 0.884 & 0.837 & 0.134 & 0.143 & 0.654 & 5.17 & 17.26 \\
     &  & ATG 2.5 & 21.57 & 0.639 & 0.862 & 0.822 & 0.165 & 0.167 & 0.584 & 5.16 & 17.81 \\
    \midrule
    \multirow{7}{*}{\shortstack[l]{Latent-\\ZOsignSGD}} & \multirow{4}{*}{60} & --- & 23.25 & 0.684 & 0.903 & 0.850 & 0.108 & 0.118 & 0.720 & 5.19 & 16.95 \\
     &  & ATG 1.5 & 22.43 & 0.664 & 0.885 & 0.838 & 0.133 & 0.143 & 0.656 & 5.18 & 17.22 \\
     &  & ATG 2.5 & 21.58 & 0.639 & 0.862 & 0.822 & 0.164 & 0.167 & 0.585 & 5.17 & 17.82 \\
     &  & TSG 60 & 23.24 & 0.683 & 0.903 & 0.850 & 0.111 & 0.122 & 0.706 & 5.29 & 16.93 \\
    \cmidrule(lr){2-12}
     & \multirow{3}{*}{120} & --- & 23.28 & 0.685 & 0.904 & 0.851 & 0.107 & 0.117 & 0.722 & 5.20 & 16.97 \\
     &  & ATG 1.5 & 22.45 & 0.665 & 0.886 & 0.838 & 0.132 & 0.142 & 0.658 & 5.20 & 17.22 \\
     &  & ATG 2.5 & 21.60 & 0.640 & 0.863 & 0.823 & 0.164 & 0.167 & 0.586 & 5.17 & 17.84 \\
    \midrule
    \multirow{7}{*}{\shortstack[l]{Latent-\\SignHunter}} & \multirow{4}{*}{60} & --- & 22.24 & 0.678 & 0.892 & 0.845 & 0.151 & 0.146 & 0.604 & 5.58 & \underline{14.92} \\
     &  & ATG 1.5 & 21.64 & 0.658 & 0.876 & 0.832 & 0.170 & 0.163 & 0.565 & 5.50 & 15.39 \\
     &  & ATG 2.5 & 20.98 & 0.634 & 0.855 & 0.818 & 0.194 & 0.182 & 0.516 & 5.44 & 16.18 \\
     &  & TSG 60 & 22.45 & 0.679 & 0.894 & 0.846 & 0.145 & 0.144 & 0.617 & 5.60 & 15.42 \\
    \cmidrule(lr){2-12}
     & \multirow{3}{*}{120} & --- & 22.33 & 0.679 & 0.893 & 0.846 & 0.148 & 0.144 & 0.612 & 5.56 & \textcolor{red}{\textbf{14.90}} \\
     &  & ATG 1.5 & 21.69 & 0.658 & 0.877 & 0.833 & 0.167 & 0.162 & 0.569 & 5.48 & 15.30 \\
     &  & ATG 2.5 & 21.02 & 0.634 & 0.856 & 0.818 & 0.192 & 0.182 & 0.519 & 5.43 & 16.16 \\
    \midrule
    \multirow{7}{*}{\shortstack[l]{Latent-\\Bandits}} & \multirow{4}{*}{60} & --- & 22.28 & 0.651 & 0.876 & 0.832 & 0.142 & 0.148 & 0.635 & \textcolor{red}{\textbf{4.90}} & 17.61 \\
     &  & ATG 1.5 & 21.61 & 0.634 & 0.859 & 0.821 & 0.166 & 0.169 & 0.584 & 4.93 & 18.13 \\
     &  & ATG 2.5 & 20.91 & 0.613 & 0.839 & 0.808 & 0.196 & 0.189 & 0.529 & 4.98 & 19.00 \\
     &  & TSG 60 & 22.46 & 0.657 & 0.881 & 0.835 & 0.138 & 0.145 & 0.641 & 5.01 & 17.45 \\
    \cmidrule(lr){2-12}
     & \multirow{3}{*}{120} & --- & 22.33 & 0.652 & 0.877 & 0.833 & 0.141 & 0.147 & 0.639 & \underline{4.90} & 17.58 \\
     &  & ATG 1.5 & 21.65 & 0.635 & 0.861 & 0.822 & 0.165 & 0.168 & 0.587 & 4.95 & 18.13 \\
     &  & ATG 2.5 & 20.94 & 0.614 & 0.840 & 0.808 & 0.194 & 0.188 & 0.531 & 4.97 & 18.91 \\
    \midrule
    \multirow{7}{*}{\shortstack[l]{Latent-\\Square}} & \multirow{4}{*}{60} & --- & 22.85 & 0.693 & 0.898 & 0.857 & 0.109 & 0.109 & 0.719 & 5.54 & 16.68 \\
     &  & ATG 1.5 & 22.12 & 0.671 & 0.881 & 0.843 & 0.134 & 0.136 & 0.657 & 5.47 & 16.86 \\
     &  & ATG 2.5 & 21.35 & 0.644 & 0.859 & 0.827 & 0.164 & 0.161 & 0.588 & 5.39 & 17.36 \\
     &  & TSG 60 & 22.93 & 0.690 & 0.899 & 0.855 & 0.113 & 0.117 & 0.705 & 5.59 & 16.82 \\
    \cmidrule(lr){2-12}
     & \multirow{3}{*}{120} & --- & 22.81 & 0.690 & 0.896 & 0.855 & 0.111 & 0.112 & 0.714 & 5.51 & 16.74 \\
     &  & ATG 1.5 & 22.08 & 0.669 & 0.879 & 0.841 & 0.136 & 0.138 & 0.652 & 5.44 & 16.83 \\
     &  & ATG 2.5 & 21.32 & 0.643 & 0.857 & 0.825 & 0.166 & 0.163 & 0.583 & 5.36 & 17.35 \\
    \midrule
    \multirow{7}{*}{\shortstack[l]{Latent-\\SimBA}} & \multirow{4}{*}{60} & --- & \textcolor{red}{\textbf{24.48}} & \textcolor{red}{\textbf{0.721}} & \textcolor{red}{\textbf{0.929}} & \textcolor{red}{\textbf{0.872}} & \textcolor{red}{\textbf{0.077}} & \textcolor{red}{\textbf{0.085}} & \textcolor{red}{\textbf{0.806}} & 5.69 & 17.09 \\
     &  & ATG 1.5 & 23.38 & 0.695 & 0.909 & 0.856 & 0.105 & 0.115 & 0.734 & 5.61 & 17.08 \\
     &  & ATG 2.5 & 22.32 & 0.665 & 0.884 & 0.837 & 0.137 & 0.144 & 0.649 & 5.53 & 17.43 \\
     &  & TSG 60 & 24.14 & 0.710 & 0.922 & 0.866 & 0.089 & 0.100 & 0.767 & 5.74 & 17.19 \\
    \cmidrule(lr){2-12}
     & \multirow{3}{*}{120} & --- & \underline{24.43} & \underline{0.719} & \underline{0.928} & \underline{0.871} & \underline{0.079} & \underline{0.086} & \underline{0.803} & 5.66 & 17.08 \\
     &  & ATG 1.5 & 23.34 & 0.694 & 0.908 & 0.855 & 0.106 & 0.116 & 0.731 & 5.58 & 17.09 \\
     &  & ATG 2.5 & 22.29 & 0.664 & 0.883 & 0.836 & 0.138 & 0.145 & 0.646 & 5.51 & 17.45 \\
    \bottomrule[1.5pt]
    \end{tabular}
  }
\end{table*}

\begin{table*}[t]
  \centering
  \small
  \caption{\textbf{Hyper-parameter values} of the latent score-based attacks.}
  \label{tab:supp-attacks}
  \begin{tabular}[t]{@{}llc@{}}
    \toprule
    Attack & Hyper-parameter & Value \\
    \midrule
    \multirow{3}{*}{General} & maximum perturbation $\varepsilon$ & 0.1 \\
     & maximum queries per active step $Q$ & 60, 120 \\
     & batch size & 1 \\
    \midrule
    \multirow{3}{*}{NES~\cite{DBLP:conf/icml/IlyasEAL18}} & \texttt{fd\_eta} & 0.05 \\
     & \texttt{lr} & 0.05 \\
     & \texttt{q} & 15 \\
    \midrule
    \multirow{3}{*}{ZOsignSGD~\cite{DBLP:conf/iclr/LiuCCH19}} & \texttt{fd\_eta} & 0.05 \\
     & \texttt{lr} & 0.05 \\
     & \texttt{q} & 30 \\
    \midrule
    SignHunter~\cite{DBLP:conf/iclr/Al-DujailiO20} & \texttt{fd\_eta} & 0.05 \\
    \bottomrule
  \end{tabular}
  \hspace{3em}
  \begin{tabular}[t]{@{}llc@{}}
    \toprule
    Attack & Hyper-parameter & Value \\
    \midrule
    \multirow{5}{*}{Bandits~\cite{DBLP:conf/iclr/IlyasEM19}} & \texttt{lr} & 0.05 \\
     & \texttt{fd\_eta} & 0.05 \\
     & \texttt{prior\_lr} & 0.005 \\
     & \texttt{prior\_size} & 28 \\
     & \texttt{prior\_exploration} & 0.1 \\
    \midrule
    Square~\cite{DBLP:conf/eccv/AndriushchenkoC20} & \texttt{p\_init} & 0.05 \\
    \midrule
    SimBA~\cite{DBLP:conf/icml/GuoGYWW19} & \texttt{delta} & 0.1 \\
    
    \bottomrule
  \end{tabular}
\end{table*}

\begin{figure*}[t]
  \centering
  \includegraphics[width=0.9\linewidth]{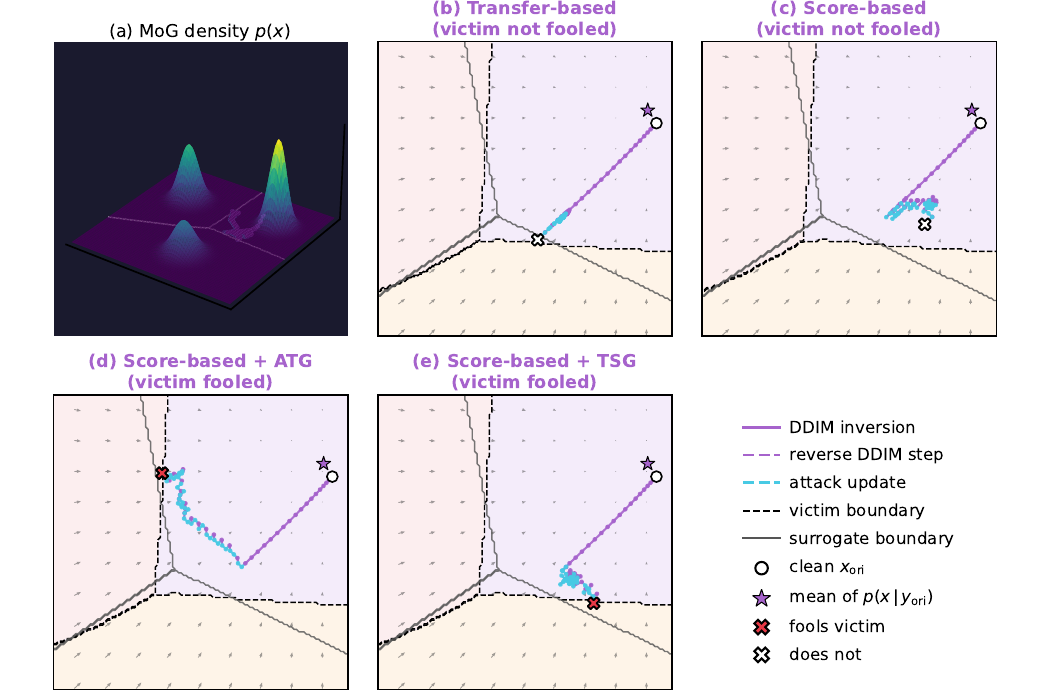}
  \caption{Simplified example on a two-dimensional mixture of three Gaussians. (a)~Density $p(x)$ with the trajectories of (b)--(e). (b)--(e)~Attacks on the same clean sample $x_{\mathrm{ori}}$ (circle), which is DDIM-inverted for 20 of 50 steps (solid) and reversed (dashed) with an attack before every step (cyan) under a per-step $L_\infty$ bound of 0.15. Background colors and dashed lines show the decision regions of the victim, grey lines the decision boundaries of the surrogate, arrows $\nabla_x\log p(x\mid y_{\mathrm{ori}})$, and the star the mean of $p(x\mid y_{\mathrm{ori}})$. The cross marks the final example and is filled when it fools the victim.}
  \label{fig:toy}
\end{figure*}

\begin{figure*}[t]
  \centering
  \includegraphics[width=0.85\textwidth]{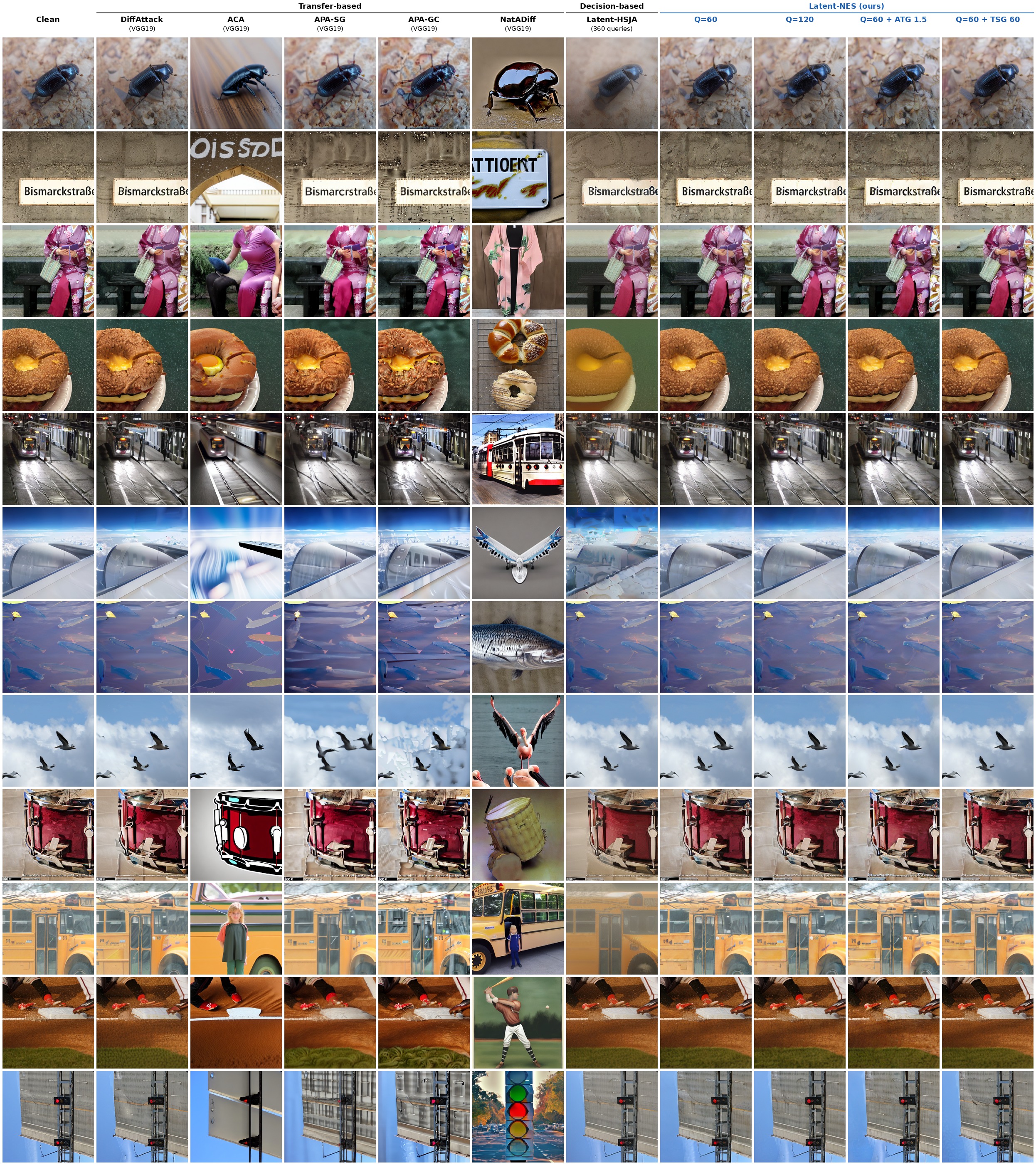}
  \caption{Extended qualitative comparison on twelve images with ResNet-50 as the victim. Transfer-based attacks use VGG19 as the surrogate, drawn at random from the eight models other than ResNet-50, and Latent-HSJA uses 360 queries per image, as Latent-NES with $Q=60$.}
  \label{fig:supp-qual}
\end{figure*}

\section{Implementation Details of Latent Score-Based \mbox{Attacks}}
\label{sec:supp-impl}
We adapt the implementations of the six score-based attacks from BlackboxBench~\cite{DBLP:journals/pami/ZhengYZCW25} to the latent space and follow its hyper-parameter settings. Tab.~\ref{tab:supp-attacks} lists the values we use, under the hyper-parameter names of BlackboxBench.

\section{Simplified Example for Visualization}
\label{sec:toy}
We use a simplified example to demonstrate the process of our methods. Figure~\ref{fig:toy} runs the attacks on a mixture of three Gaussians, for which the noise prediction of each class is available in closed form. The victim is an MLP and the surrogate a linear classifier, both trained on samples of the mixture. A clean sample of $y_{\mathrm{ori}}$ is DDIM-inverted and then reversed with an untargeted attack on $\hat{x}_{0|t}$ before every step and once after the last step, in the same order as our attack on images. The transfer-based attack follows the gradient of the surrogate and crosses the decision boundary of the surrogate but not that of the victim. The score-based attack estimates the gradient of the victim with NES from 8 queries per step, but every DDIM step pulls the sample back towards $p(x\mid y_{\mathrm{ori}})$, so it stays in the region of $y_{\mathrm{ori}}$. ATG with $s=0.5$ towards the runner-up class and TSG, which searches the condition with 8 queries per step, shift the reverse process itself, and both cross the decision boundary of the victim. Over 100 clean samples, the victim is fooled in 29\%, 10\%, 100\% and 49\% of the runs of the transfer-based attack, the score-based attack, ATG and TSG, respectively.

\section{Qualitative Results}
\label{sec:supp-qual}
Figure~\ref{fig:supp-qual} compares the adversarial examples of the selected attacks on twelve images of the ImageNet-Compatible dataset in the black-box setting. ACA replaces large parts of the content, and NatADiff synthesizes a new image of the class without a clean reference. APA-SG and APA-GC add visible artifacts, and Latent-HSJA leaves traces of its starting image of another class. DiffAttack and Latent-NES preserve the content and layout of the clean images, and the guidance of Latent-NES changes only fine textures.

%% file: main.bib
@String(ECCV= {Eur. Conf. Comput. Vis.})

@String(NIPS= {Adv. Neural Inform. Process. Syst.})

@String(ECCV  = {ECCV})

@String(NIPS  = {NeurIPS})

@inproceedings{DBLP:journals/corr/SzegedyZSBEGF13,
  author       = {Christian Szegedy and
                  Wojciech Zaremba and
                  Ilya Sutskever and
                  Joan Bruna and
                  Dumitru Erhan and
                  Ian J. Goodfellow and
                  Rob Fergus},
  title        = {Intriguing properties of neural networks},
  booktitle    = {International Conference on Learning Representations},
  year         = 2014,
}

@inproceedings{DBLP:journals/corr/GoodfellowSS14,
  author       = {Ian J. Goodfellow and
                  Jonathon Shlens and
                  Christian Szegedy},
  title        = {Explaining and Harnessing Adversarial Examples},
  booktitle    = {International Conference on Learning Representations},
  year         = 2015,
}

@inproceedings{DBLP:conf/iclr/MadryMSTV18,
  author       = {Aleksander Madry and
                  Aleksandar Makelov and
                  Ludwig Schmidt and
                  Dimitris Tsipras and
                  Adrian Vladu},
  title        = {Towards Deep Learning Models Resistant to Adversarial Attacks},
  booktitle    = {International Conference on Learning Representations},
  year         = 2018,
}

@inproceedings{DBLP:conf/sp/Carlini017,
  author       = {Nicholas Carlini and
                  David A. Wagner},
  title        = {Towards Evaluating the Robustness of Neural Networks},
  booktitle    = {Proceedings of the {IEEE} Symposium on Security and Privacy},
  pages        = {39--57},
  publisher    = {{IEEE} Computer Society},
  year         = 2017,
}

@inproceedings{DBLP:conf/eurosp/PapernotMJFCS16,
  author       = {Nicolas Papernot and
                  Patrick D. McDaniel and
                  Somesh Jha and
                  Matt Fredrikson and
                  Z. Berkay Celik and
                  Ananthram Swami},
  title        = {The Limitations of Deep Learning in Adversarial Settings},
  booktitle    = {Proceedings of the {IEEE} European Symposium on Security and Privacy},
  pages        = {372--387},
  publisher    = {{IEEE}},
  year         = {2016},
}

@inproceedings{DBLP:conf/cvpr/Moosavi-Dezfooli16,
  author       = {Seyed{-}Mohsen Moosavi{-}Dezfooli and
                  Alhussein Fawzi and
                  Pascal Frossard},
  title        = {DeepFool: {A} Simple and Accurate Method to Fool Deep Neural Networks},
  booktitle    = {Proceedings of the {IEEE} Conference on Computer Vision and Pattern Recognition},
  pages        = {2574--2582},
  publisher    = {{IEEE} Computer Society},
  year         = {2016},
}

@inproceedings{DBLP:conf/iclr/KurakinGB17a,
  author       = {Alexey Kurakin and
                  Ian J. Goodfellow and
                  Samy Bengio},
  title        = {Adversarial examples in the physical world},
  booktitle    = {{International Conference on Learning Representations} Workshop Track Proceedings},
  year         = {2017},
}

@article{DBLP:journals/access/AkhtarM18,
  author       = {Naveed Akhtar and
                  Ajmal S. Mian},
  title        = {Threat of Adversarial Attacks on Deep Learning in Computer Vision:
                  {A} Survey},
  journal      = {{IEEE} Access},
  volume       = {6},
  pages        = {14410--14430},
  year         = {2018},
}

@inproceedings{DBLP:conf/nips/SongSKE18,
  author       = {Yang Song and
                  Rui Shu and
                  Nate Kushman and
                  Stefano Ermon},
  title        = {Constructing Unrestricted Adversarial Examples with Generative Models},
  booktitle    = {Proceedings of the Advances in Neural Information Processing Systems},
  pages        = {8322--8333},
  year         = {2018},
}

@article{DBLP:journals/pami/ChenCCZZS25,
  author       = {Jianqi Chen and
                  Hao Chen and
                  Keyan Chen and
                  Yilan Zhang and
                  Zhengxia Zou and
                  Zhenwei Shi},
  title        = {Diffusion Models for Imperceptible and Transferable Adversarial Attack},
  journal      = {{IEEE} Trans. Pattern Anal. Mach. Intell.},
  volume       = {47},
  number       = {2},
  pages        = {961--977},
  year         = {2025},
}

@inproceedings{DBLP:conf/icml/ZhangYJXGJ19,
  author       = {Hongyang Zhang and
                  Yaodong Yu and
                  Jiantao Jiao and
                  Eric P. Xing and
                  Laurent El Ghaoui and
                  Michael I. Jordan},
  title        = {Theoretically Principled Trade-off between Robustness and Accuracy},
  booktitle    = {Proceedings of the International Conference on Machine Learning},
  volume       = {97},
  pages        = {7472--7482},
  publisher    = {{PMLR}},
  year         = {2019},
}

@inproceedings{shi2021online,
  title={Online adversarial purification based on self-supervision},
  author={Shi, Changhao and Holtz, Chester and Mishne, Gal},
  booktitle={International Conference on Learning Representations},
  year={2021}
}

@inproceedings{DBLP:conf/icml/OdenaOS17,
  author       = {Augustus Odena and
                  Christopher Olah and
                  Jonathon Shlens},
  title        = {Conditional Image Synthesis with Auxiliary Classifier GANs},
  booktitle    = {Proceedings of the International Conference on Machine Learning},
  volume       = {70},
  pages        = {2642--2651},
  publisher    = {{PMLR}},
  year         = {2017},
}

@inproceedings{DBLP:conf/iclr/BhattadCLLF20,
  author       = {Anand Bhattad and
                  Min Jin Chong and
                  Kaizhao Liang and
                  Bo Li and
                  David A. Forsyth},
  title        = {Unrestricted Adversarial Examples via Semantic Manipulation},
  booktitle    = {Proceedings of the International Conference on Learning Representations},
  publisher    = {OpenReview.net},
  year         = {2020},
}

@inproceedings{DBLP:conf/cvpr/HosseiniP18,
  author       = {Hossein Hosseini and
                  Radha Poovendran},
  title        = {Semantic Adversarial Examples},
  booktitle    = {Proceedings of the {IEEE} Conference on Computer Vision and Pattern Recognition
                  Workshops},
  pages        = {1614--1619},
  publisher    = {Computer Vision Foundation / {IEEE} Computer Society},
  year         = {2018},
}

@inproceedings{DBLP:conf/cvpr/ShamsabadiSC20,
  author       = {Ali Shahin Shamsabadi and
                  Ricardo S{\'{a}}nchez{-}Matilla and
                  Andrea Cavallaro},
  title        = {ColorFool: Semantic Adversarial Colorization},
  booktitle    = {Proceedings of the {IEEE/CVF} Conference on Computer Vision and Pattern Recognition},
  pages        = {1148--1157},
  publisher    = {Computer Vision Foundation / {IEEE}},
  year         = {2020},
}

@inproceedings{DBLP:conf/bmvc/ZhaoLL20,
  author       = {Zhengyu Zhao and
                  Zhuoran Liu and
                  Martha A. Larson},
  title        = {Adversarial Color Enhancement: Generating Unrestricted Adversarial Images by Optimizing a Color Filter},
  booktitle    = {Proceedings of the British Machine Vision Conference},
  publisher    = {{BMVA} Press},
  year         = {2020},
}

@inproceedings{DBLP:conf/nips/ChenLWJDZ23,
  author       = {Zhaoyu Chen and
                  Bo Li and
                  Shuang Wu and
                  Kaixun Jiang and
                  Shouhong Ding and
                  Wenqiang Zhang},
  title        = {Content-based Unrestricted Adversarial Attack},
  booktitle    = {Proceedings of the Advances in Neural Information Processing Systems},
  year         = {2023},
}

@inproceedings{DBLP:conf/nips/YuanZGCS22,
  author       = {Shengming Yuan and
                  Qilong Zhang and
                  Lianli Gao and
                  Yaya Cheng and
                  Jingkuan Song},
  title        = {Natural Color Fool: Towards Boosting Black-box Unrestricted Attacks},
  booktitle    = {Proceedings of the Advances in Neural Information Processing Systems},
  year         = {2022},
}

@inproceedings{DBLP:conf/iclr/XiaoZ0HLS18,
  author       = {Chaowei Xiao and
                  Jun{-}Yan Zhu and
                  Bo Li and
                  Warren He and
                  Mingyan Liu and
                  Dawn Song},
  title        = {Spatially Transformed Adversarial Examples},
  booktitle    = {Proceedings of the International Conference on Learning Representations},
  publisher    = {OpenReview.net},
  year         = {2018},
}

@inproceedings{DBLP:conf/cvpr/RombachBLEO22,
  author       = {Robin Rombach and
                  Andreas Blattmann and
                  Dominik Lorenz and
                  Patrick Esser and
                  Bj{\"{o}}rn Ommer},
  title        = {High-Resolution Image Synthesis with Latent Diffusion Models},
  booktitle    = {Proceedings of the {IEEE/CVF} Conference on Computer Vision and Pattern Recognition},
  pages        = {10674--10685},
  publisher    = {{IEEE}},
  year         = {2022},
}

@inproceedings{DBLP:conf/eccv/DaiLX24,
  author       = {Xuelong Dai and
                  Kaisheng Liang and
                  Bin Xiao},
  title        = {AdvDiff: Generating Unrestricted Adversarial Examples Using Diffusion
                  Models},
  booktitle    = {Proceedings of the European Conference on Computer Vision},
  volume       = {15104},
  pages        = {93--109},
  publisher    = {Springer},
  year         = {2024},
}

@inproceedings{DBLP:conf/iccv/ChenGZY023,
  author       = {Xinquan Chen and
                  Xitong Gao and
                  Juanjuan Zhao and
                  Kejiang Ye and
                  Cheng{-}Zhong Xu},
  title        = {AdvDiffuser: Natural Adversarial Example Synthesis with Diffusion
                  Models},
  booktitle    = {Proceedings of the International Conference on Computer Vision},
  pages        = {4539--4549},
  publisher    = {{IEEE}},
  year         = {2023},
}

@inproceedings{DBLP:conf/ccs/PapernotMGJCS17,
  author       = {Nicolas Papernot and
                  Patrick D. McDaniel and
                  Ian J. Goodfellow and
                  Somesh Jha and
                  Z. Berkay Celik and
                  Ananthram Swami},
  title        = {Practical Black-Box Attacks against Machine Learning},
  booktitle    = {Proceedings of the {ACM} on Asia Conference on Computer and Communications Security},
  pages        = {506--519},
  publisher    = {{ACM}},
  year         = {2017},
}

@inproceedings{DBLP:conf/nips/0001ZW24,
  author       = {Andi Zhang and
                  Mingtian Zhang and
                  Damon Wischik},
  title        = {Constructing Semantics-Aware Adversarial Examples with a Probabilistic
                  Perspective},
  booktitle    = {Proceedings of the Advances in Neural Information Processing Systems},
  year         = {2024},
}

@inproceedings{DBLP:conf/iclr/CollinsVFM26,
  author       = {Max Collins and
                  Jordan Vice and
                  Tim French and
                  Ajmal Mian},
  editor       = {Carl Vondrick and
                  Bharath Hariharan and
                  Colin Raffel and
                  Lerrel Pinto and
                  Diyi Yang and
                  Aleksandra Faust},
  title        = {NatADiff: Adversarial Boundary Guidance for Natural Adversarial Diffusion},
  booktitle    = {Proceedings of the International Conference on Learning Representations},
  publisher    = {proceedings.iclr.cc},
  year         = {2026},
}

@inproceedings{DBLP:conf/iclr/ZhangDMK26,
  author       = {Andi Zhang and
                  Xuan Ding and
                  Steven McDonagh and
                  Samuel Kaski},
  editor       = {Carl Vondrick and
                  Bharath Hariharan and
                  Colin Raffel and
                  Lerrel Pinto and
                  Diyi Yang and
                  Aleksandra Faust},
  title        = {Concept-based Adversarial Attack: a Probabilistic Perspective},
  booktitle    = {Proceedings of the International Conference on Learning Representations},
  publisher    = {proceedings.iclr.cc},
  year         = {2026},
}

@inproceedings{DBLP:conf/nips/HoJA20,
  author       = {Jonathan Ho and
                  Ajay Jain and
                  Pieter Abbeel},
  title        = {Denoising Diffusion Probabilistic Models},
  booktitle    = {Proceedings of the Advances in Neural Information Processing Systems},
  year         = {2020},
}

@article{DBLP:journals/pami/ZhengYZCW25,
  author       = {Meixi Zheng and
                  Xuanchen Yan and
                  Zihao Zhu and
                  Hongrui Chen and
                  Baoyuan Wu},
  title        = {BlackboxBench: {A} Comprehensive Benchmark of Black-Box Adversarial
                  Attacks},
  journal      = {{IEEE} Trans. Pattern Anal. Mach. Intell.},
  volume       = {47},
  number       = {9},
  pages        = {7867--7885},
  year         = {2025},
}

@inproceedings{DBLP:conf/icml/GuoGYWW19,
  author       = {Chuan Guo and
                  Jacob R. Gardner and
                  Yurong You and
                  Andrew Gordon Wilson and
                  Kilian Q. Weinberger},
  title        = {Simple Black-box Adversarial Attacks},
  booktitle    = {Proceedings of the International Conference on Machine Learning},
  volume       = {97},
  pages        = {2484--2493},
  publisher    = {{PMLR}},
  year         = {2019},
}

@inproceedings{DBLP:conf/eccv/AndriushchenkoC20,
  author       = {Maksym Andriushchenko and
                  Francesco Croce and
                  Nicolas Flammarion and
                  Matthias Hein},
  title        = {Square Attack: {A} Query-Efficient Black-Box Adversarial Attack via
                  Random Search},
  booktitle    = {Proceedings of the European Conference on Computer Vision},
  volume       = {12368},
  pages        = {484--501},
  publisher    = {Springer},
  year         = {2020},
}

@inproceedings{DBLP:conf/icml/IlyasEAL18,
  author       = {Andrew Ilyas and
                  Logan Engstrom and
                  Anish Athalye and
                  Jessy Lin},
  title        = {Black-box Adversarial Attacks with Limited Queries and Information},
  booktitle    = {Proceedings of the International Conference on Machine Learning},  
  volume       = {80},
  pages        = {2142--2151},
  publisher    = {{PMLR}},
  year         = {2018},
}

@inproceedings{DBLP:conf/iclr/IlyasEM19,
  author       = {Andrew Ilyas and
                  Logan Engstrom and
                  Aleksander Madry},
  title        = {Prior Convictions: Black-box Adversarial Attacks with Bandits and
                  Priors},
  booktitle    = {Proceedings of the International Conference on Learning Representations},
  publisher    = {OpenReview.net},
  year         = {2019},
}

@inproceedings{DBLP:conf/iclr/LiuCCH19,
  author       = {Sijia Liu and
                  Pin{-}Yu Chen and
                  Xiangyi Chen and
                  Mingyi Hong},
  title        = {signSGD via Zeroth-Order Oracle},
  booktitle    = {Proceedings of the International Conference on Learning Representations},
  publisher    = {OpenReview.net},
  year         = {2019},
}

@inproceedings{DBLP:conf/iclr/Al-DujailiO20,
  author       = {Abdullah Al{-}Dujaili and
                  Una{-}May O'Reilly},
  title        = {Sign Bits Are All You Need for Black-Box Attacks},
  booktitle    = {Proceedings of the International Conference on Learning Representations},
  publisher    = {OpenReview.net},
  year         = {2020},
}

@inproceedings{DBLP:conf/iclr/SongME21,
  author       = {Jiaming Song and
                  Chenlin Meng and
                  Stefano Ermon},
  title        = {Denoising Diffusion Implicit Models},
  booktitle    = {Proceedings of the International Conference on Learning Representations},
  publisher    = {OpenReview.net},
  year         = {2021},
}

@article{efron2011tweedie,
  title={Tweedie’s formula and selection bias},
  author={Efron, Bradley},
  journal={Journal of the American Statistical Association},
  volume={106},
  number={496},
  pages={1602--1614},
  year={2011},
  publisher={Taylor \& Francis}
}

@article{DBLP:journals/corr/abs-2207-12598,
  author       = {Jonathan Ho and
                  Tim Salimans},
  title        = {Classifier-Free Diffusion Guidance},
  journal      = {CoRR},
  volume       = {abs/2207.12598},
  year         = {2022},
  doi          = {10.48550/ARXIV.2207.12598},
  eprinttype   = {arXiv},
  eprint       = {2207.12598},
}

@inproceedings{DBLP:conf/eccv/NaJK22,
  author       = {Dongbin Na and
                  Sangwoo Ji and
                  Jong Kim},
  title        = {Unrestricted Black-Box Adversarial Attack Using {GAN} with Limited
                  Queries},
  booktitle    = { {ECCV} 2022 Workshops },
  volume       = {13801},
  pages        = {467--482},
  publisher    = {Springer},
  year         = {2022},
}

@inproceedings{wang2026decision,
  title={Decision-Only Adversarial Editing with Diffusion Models},
  author={Wang, Yaowen and Cullina, Daniel},
  booktitle={Proceedings of the IEEE/CVF Conference on Computer Vision and Pattern Recognition Workshops},
  pages={1451--1460},
  year={2026}
}

@inproceedings{DBLP:conf/nips/JiangCGLFGTLZ25,
  author       = {Kaixun Jiang and
                  Zhaoyu Chen and
                  Haijing Guo and
                  Jinglun Li and
                  Jiyuan Fu and
                  Pinxue Guo and
                  Hao Tang and
                  Bo Li and
                  Wenqiang Zhang},
  title        = {Enhancing Diffusion-based Unrestricted Adversarial Attacks via Adversary
                  Preferences Alignment},
  booktitle    = NIPS,
  year         = {2025},
}

@article{DBLP:journals/corr/abs-1804-00097,
  author       = {Alexey Kurakin and
                  Ian J. Goodfellow and
                  Samy Bengio and
                  Yinpeng Dong and
                  Fangzhou Liao and
                  Ming Liang and
                  Tianyu Pang and
                  Jun Zhu and
                  Xiaolin Hu and
                  Cihang Xie and
                  Jianyu Wang and
                  Zhishuai Zhang and
                  Zhou Ren and
                  Alan L. Yuille and
                  Sangxia Huang and
                  Yao Zhao and
                  Yuzhe Zhao and
                  Zhonglin Han and
                  Junjiajia Long and
                  Yerkebulan Berdibekov and
                  Takuya Akiba and
                  Seiya Tokui and
                  Motoki Abe},
  title        = {Adversarial Attacks and Defences Competition},
  journal      = {CoRR},
  volume       = {abs/1804.00097},
  year         = {2018},
}

@inproceedings{DBLP:conf/nips/HeuselRUNH17,
  author    = {Martin Heusel and Hubert Ramsauer and Thomas Unterthiner and Bernhard Nessler and Sepp Hochreiter},
  title     = {{GANs} Trained by a Two Time-Scale Update Rule Converge to a Local {Nash} Equilibrium},
  booktitle = {Advances in Neural Information Processing Systems},
  pages     = {6626--6637},
  year      = {2017}
}

@inproceedings{DBLP:conf/cvpr/DongLPS0HL18,
  author       = {Yinpeng Dong and
                  Fangzhou Liao and
                  Tianyu Pang and
                  Hang Su and
                  Jun Zhu and
                  Xiaolin Hu and
                  Jianguo Li},
  title        = {Boosting Adversarial Attacks With Momentum},
  booktitle    = {Proceedings of the {IEEE} Conference on Computer Vision and Pattern Recognition},
  publisher    = {Computer Vision Foundation / {IEEE} Computer Society},
  year         = {2018},
}

@inproceedings{DBLP:conf/cvpr/XieZZBWRY19,
  author       = {Cihang Xie and
                  Zhishuai Zhang and
                  Yuyin Zhou and
                  Song Bai and
                  Jianyu Wang and
                  Zhou Ren and
                  Alan L. Yuille},
  title        = {Improving Transferability of Adversarial Examples With Input Diversity},
  booktitle    = {Proceedings of the {IEEE} Conference on Computer Vision and Pattern Recognition},
  pages        = {2730--2739},
  publisher    = {Computer Vision Foundation / {IEEE}},
  year         = {2019},
}

@inproceedings{DBLP:conf/cvpr/DongPSZ19,
  author       = {Yinpeng Dong and
                  Tianyu Pang and
                  Hang Su and
                  Jun Zhu},
  title        = {Evading Defenses to Transferable Adversarial Examples by Translation-Invariant
                  Attacks},
  booktitle    = {Proceedings of the {IEEE} Conference on Computer Vision and Pattern Recognition},
  pages        = {4312--4321},
  publisher    = {Computer Vision Foundation / {IEEE}},
  year         = {2019},
}

@inproceedings{DBLP:conf/eccv/GaoZSLS20,
  author       = {Lianli Gao and
                  Qilong Zhang and
                  Jingkuan Song and
                  Xianglong Liu and
                  Heng Tao Shen},
  title        = {Patch-Wise Attack for Fooling Deep Neural Network},
  booktitle    = {Proceedings of the European Conference on Computer Vision},
  volume       = {12373},
  pages        = {307--322},
  publisher    = {Springer},
  year         = {2020},
}

@inproceedings{DBLP:conf/eccv/LongZZGLZS22,
  author       = {Yuyang Long and
                  Qilong Zhang and
                  Boheng Zeng and
                  Lianli Gao and
                  Xianglong Liu and
                  Jian Zhang and
                  Jingkuan Song},
  title        = {Frequency Domain Model Augmentation for Adversarial Attack},
  booktitle    = {Proceedings of the European Conference on Computer Vision},
  volume       = {13664},
  pages        = {549--566},
  publisher    = {Springer},
  year         = {2022},
}

@inproceedings{DBLP:conf/iclr/TramerKPGBM18,
  author       = {Florian Tram{\`{e}}r and
                  Alexey Kurakin and
                  Nicolas Papernot and
                  Ian J. Goodfellow and
                  Dan Boneh and
                  Patrick D. McDaniel},
  title        = {Ensemble Adversarial Training: Attacks and Defenses},
  booktitle    = {Proceedings of the International Conference on Learning Representations },
  publisher    = {OpenReview.net},
  year         = {2018},
}

@inproceedings{DBLP:conf/iclr/BrendelRB18,
  author       = {Wieland Brendel and
                  Jonas Rauber and
                  Matthias Bethge},
  title        = {Decision-Based Adversarial Attacks: Reliable Attacks Against Black-Box
                  Machine Learning Models},
  booktitle    = {Proceedings of the International Conference on Learning Representations},
  publisher    = {OpenReview.net},
  year         = {2018},
}

@inproceedings{DBLP:conf/sp/ChenJW20,
  author       = {Jianbo Chen and
                  Michael I. Jordan and
                  Martin J. Wainwright},
  title        = {HopSkipJumpAttack: {A} Query-Efficient Decision-Based Attack},
  booktitle    = {Proceedings of the {IEEE} Symposium on Security and Privacy},
  pages        = {1277--1294},
  publisher    = {{IEEE}},
  year         = {2020},
}

@inproceedings{DBLP:conf/nips/DolatabadiEL20,
  author       = {Hadi Mohaghegh Dolatabadi and
                  Sarah M. Erfani and
                  Christopher Leckie},
  title        = {AdvFlow: Inconspicuous Black-box Adversarial Attacks using Normalizing
                  Flows},
  booktitle    = {Proceedings of the Advances in Neural Information Processing Systems},
  year         = {2020},
}

@inproceedings{DBLP:conf/nips/XueA0C23,
  author       = {Haotian Xue and
                  Alexandre Araujo and
                  Bin Hu and
                  Yongxin Chen},
  editor       = {Alice Oh and
                  Tristan Naumann and
                  Amir Globerson and
                  Kate Saenko and
                  Moritz Hardt and
                  Sergey Levine},
  title        = {Diffusion-Based Adversarial Sample Generation for Improved Stealthiness
                  and Controllability},
  booktitle    = {Proceedings of the Advances in Neural Information Processing Systems},
  year         = {2023},
}

@article{DBLP:journals/neco/Hinton02,
  author       = {Geoffrey E. Hinton},
  title        = {Training Products of Experts by Minimizing Contrastive Divergence},
  journal      = {Neural Comput.},
  volume       = {14},
  number       = {8},
  pages        = {1771--1800},
  year         = {2002},
}

@article{DBLP:journals/tip/ZhuCLC00ZCH22,
  author       = {Yao Zhu and
                  Yuefeng Chen and
                  Xiaodan Li and
                  Kejiang Chen and
                  Yuan He and
                  Xiang Tian and
                  Bolun Zheng and
                  Yaowu Chen and
                  Qingming Huang},
  title        = {Toward Understanding and Boosting Adversarial Transferability From
                  a Distribution Perspective},
  journal      = {{IEEE} Trans. Image Process.},
  volume       = {31},
  pages        = {6487--6501},
  year         = {2022},
}
